\pdfoutput=1

\documentclass[11pt]{article}

\usepackage[final]{coling}

\usepackage{times}
\usepackage{multirow}
\usepackage{longtable}
\usepackage{latexsym}
\usepackage{amsmath}
\usepackage{subcaption}
\usepackage{dirtytalk}
\usepackage{float}
\usepackage{hyperref}
\usepackage{inconsolata}
\usepackage{color,soul}

\usepackage{listings}
\usepackage[T1]{fontenc}

\usepackage[utf8]{inputenc}

\usepackage{microtype}

\usepackage{inconsolata}

\usepackage{graphicx}

\title{Bridging Context Gaps: Enhancing Comprehension in Long-Form Social Conversations Through Contextualized Excerpts}

\author{Shrestha Mohanty$^{\dagger}$, Sarah Xuan$^{\dagger}$, Jacob Jobraeel$^{\dagger}$, Anurag Kumar$^{\ddagger}$, Deb Roy$^{\dagger}$, Jad Kabbara$^{\dagger}$\\
  $^{\dagger }$ Massachusetts Institute of Technology, 
  $^{\ddagger}$ Meta\\
  \texttt{shresmoh@mit.edu}}
    
\begin{document}
\maketitle
\begin{abstract}


We focus on enhancing comprehension in small-group recorded conversations, which serve as a medium to bring people together and provide a space for sharing personal stories and experiences on crucial social matters. One way to parse and convey information from these conversations is by sharing highlighted excerpts in subsequent conversations. This can help promote a collective understanding of relevant issues, by highlighting perspectives and experiences to other groups of people who might otherwise be unfamiliar with and thus unable to relate to these experiences. 
The primary challenge that arises then is that excerpts taken from one conversation and shared in another setting might be missing crucial context or key elements that were previously introduced in the original conversation. This problem is exacerbated when conversations become lengthier and richer in themes and shared experiences. To address this, we explore how Large Language Models (LLMs) can enrich these excerpts by providing socially relevant context. We present approaches for effective contextualization to improve comprehension, readability, and empathy. We show significant improvements in understanding, as assessed through subjective and objective evaluations. While LLMs can offer valuable context, they struggle with capturing key social aspects. We release the Human-annotated Salient Excerpts (HSE) dataset to support future work. Additionally, we show how context-enriched excerpts can provide more focused and comprehensive conversation summaries.
\end{abstract}

\section{Introduction}
\label{sec:introduction}

\begin{figure*}[h!]
    \centering
    \includegraphics[width=0.9\textwidth]{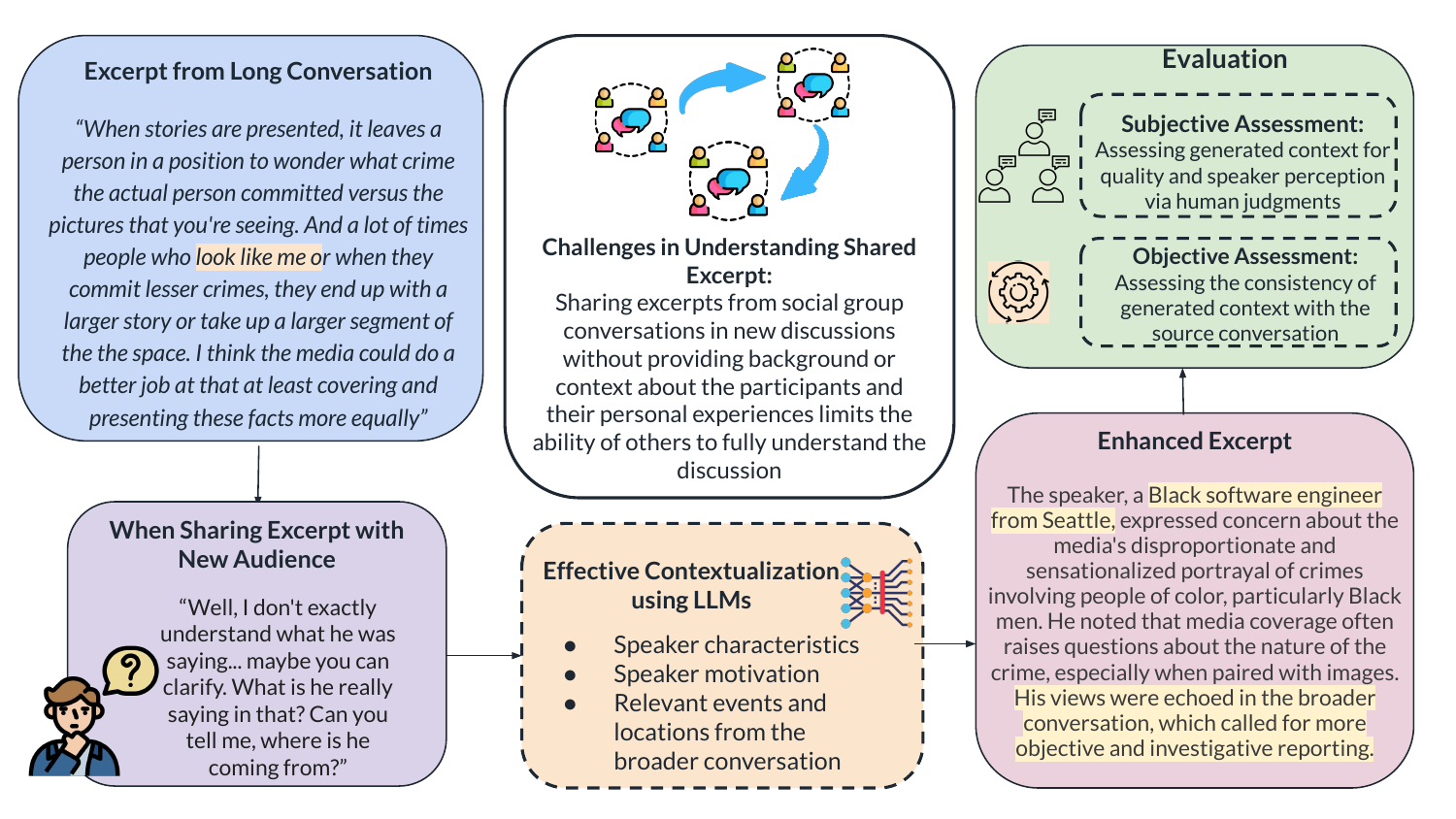}
    \caption{\textbf{Overview of Effective Contextualization of Salient Excerpts in Social Conversations:} The figure illustrates the challenges posed when excerpts from longer social group conversations are shared without sufficient context. To address this, a large language model (LLM) is employed to generate context that incorporates key social attributes, aiming to enhance the comprehension of the excerpt, a process we term effective contextualization. This enhanced context is then evaluated through human judgment to gauge its usefulness and through objective measures of faithfulness to assess its accuracy.}
    \label{fig:overview}
\end{figure*}

In recent years, there has been a troubling rise in polarization, both on social media and even in public spaces in society, with these spaces being increasingly dominated by loud and extreme voices. Motivated by this reality, we focus on small-group conversations as a medium to bring people together, strengthen community building and understanding, and present people a space to share personal stories and experiences related to crucial matters that affect their daily lives. These conversations can be recorded (with consent) thus allowing the sharing of excerpts from one conversation in many other conversations. This can help in promoting a collective understanding of relevant issues, highlighting perspectives and experiences to other groups of people who might otherwise be unfamiliar with, and thus unable to relate to, these experiences.

A key challenge here is that excerpts taken from one conversation and shared in another setting might be missing key elements or crucial context that was previously introduced in the original conversation, e.g., background information about the speaker, their intent, or key details about the story or perspective being shared. This could lead to misunderstandings or misinterpretation and could elicit negative reactions from the receiving side as they may not fully grasp the background that shapes the speaker's words. This issue is increasingly common in today's digital landscape, where missing context often fuels misunderstandings \cite{mauranen2006signaling} and perpetuates false and harmful stereotypes that contribute to dehumanization \cite{Roy2023}. Moreover, this problem is exacerbated when, as in our case, conversations become longer and richer in themes and shared experiences.

In this work, we examine the role of highlighted salient excerpts and the influence of social factors in understanding long-form group conversations, spanning tens of thousands of words. We assess how well these excerpts can convey socially relevant information and explore the effectiveness of contextualizing excerpts to improve readability and overall understanding, a process we term \emph{effective contextualization} of the excerpt. More specifically, we examine how Large Language Models (LLMs) can retrieve and synthesize additional information related to social attributes to augment the excerpts from lengthy social conversations.

The conversations~\cite{schroeder-etal-2024-fora} we focus on involve diverse speakers discussing and sharing personal stories on topics that include public health, affordable housing, and neighborhood safety among others. While LLMs have been effective in synthesizing information in structured settings like news or meetings~\cite{goyal2022news, feigenblat2021tweetsumm, laskar2023building}, their application to social conversations is less studied. 

Figure \ref{fig:overview} illustrates the novel problem we examine in this work. It presents a human-highlighted excerpt from a long group conversation about systemic bias in media representation. The excerpt, itself, without additional context, is difficult to interpret. Key elements such as the speaker's identity, their intent, and the broader conversational topic remain unclear. To address this, we employ a large language model (LLM) for effective contextualization with relevant social attributes, improving the excerpt’s comprehensibility. We develop two approaches for effective contextualization of a given excerpt from a conversation. The first one relies on the zero-shot \cite{kojima2022large} capabilities of LLMs whereas, for the second one, we instruct the LLM to explicitly focus on socially grounded attributes to effectively contextualize the excerpt. 
The quality of the enhanced excerpt is then assessed through both human judgment for usefulness and objective measures of faithfulness for accuracy.

The key contributions of this work are:  
\begin{enumerate}  
    \item We explore the use of LLMs for the analysis and understanding of long-form social group conversations through \emph{highlighted salient excerpts}. To the best of our knowledge, this is the first work to explore the use of salient excerpts for understanding long multi-party social conversations. These conversations present unique challenges due to their distinctive characteristics, such as personal storytelling, social cues, topic shifts, informal language, and diverse perspectives. While our method is tailored for social conversations, it can also be adapted to other domains, such as structured meetings.  
    
    \item We introduce a new dataset, \emph{Human-annotated Salient Excerpts} (HSE), built on the Fora dataset~\cite{schroeder-etal-2024-fora}, to enable exploration of this novel problem.  
    
    \item We introduce the concept of \emph{effective contextualization}, exploring two approaches for enriching salient excerpts with social context using LLMs. Our analysis shows that while LLMs can synthesize useful context, they struggle to accurately extract key social information.
    
    \item We demonstrate the utility of salient excerpts and their contextualized versions for summarizing lengthy social conversations. Summaries based on effectively contextualized excerpts are rated higher in subjective evaluations, underscoring the value of integrating social attributes when designing automated methods for understanding social group conversations.  
\end{enumerate}

\section{Related Work}
\label{sec:rel-work}

\noindent \textbf{Social dialogue:} Social group conversations, in the context of our study, refer to discussions within small groups where participants share experiences, perspectives, and viewpoints on topics of communal relevance. These conversations are rich in social dynamics, often featuring personal stories, emotional expressions, and diverse perspectives. They have been recognized as valuable across various fields, offering insights into collective understanding, civic governance, and nuanced perspectives in the social sciences \cite{Roy2023, schroeder-etal-2024-fora}. Past work highlighted the importance of such dialogues in fostering better decision-making and legitimizing democratic outcomes.  Dialogue networks \cite{Roy2023} rely on such conversations among a small group of people to capture their perspectives on local issues and enable participation in civic and democratic processes. Deliberative polling has demonstrated the role of discourse in improving decision outcomes \cite{fishkin1997voice}, while storytelling has been shown to enhance productive deliberation in forums \cite{ryfe2006narrative}. Sustained dialogue, as noted by \cite{saunders1999public}, plays a critical role in transforming relationships amidst deeply rooted conflicts. This body of work underscores the significance of social group conversations as a medium for understanding collective reasoning, resolving conflicts, and studying broader societal dynamics. These conversations can be lengthy, on average over an hour long \cite{schroeder-etal-2024-fora} necessitating approaches to distill key points.

\noindent \textbf{Socially aware NLP models:} Recent advances in NLP and LLMs have enabled the development of more socially aware models. \citet{ziems2024can} discuss the growing ability to create such models for computational social science problems. Several works \cite{flek2020returning,hovy2021importance,yang2024call} emphasized the importance of understanding the relationship between language and social context, advocating for the development of socially aware NLP systems. Our work builds on these efforts by integrating attention mechanisms to better capture social factors within conversations.

\noindent \textbf{Extracting social dynamics and speaker characteristics in NLP:} LLMs have been applied to understand both emotional undertones and social meaning in conversations, as well as to extract speaker characteristics. \citet{dutt2024leveragingmachinegeneratedrationalesfacilitate} use machine-generated rationales to deduce the emotional and social meaning behind conversational statements, while \cite{chae2023large} focuses on extracting social stances from online dialogue. Additionally, \cite{jurafsky2009extracting} and \cite{broniatowski2012extracting} presented methods for identifying the speaker's personality and identity from the text. We are inspired by these methods to explore the use of LLM-generated enriched contexts to enhance the understanding of both the speaker’s identity and the social dynamics surrounding the conversation excerpt.

\noindent \textbf{Human-in-the-loop for contextualization and synthesis:} Human-in-the-loop methods have proven effective in refining LLM-generated content. \cite{jiang-etal-2024-leveraging} involves human experts to improve the quality of LLM-generated content used to help in the understanding of complex legal concepts. Additionally, \cite{chen-etal-2023-human} illustrates how human-highlighted information can improve LLM-generated summaries while 
\cite{yao2023improvingsummarizationhumanedits} 
shows enhanced domain-specific summarization by incorporating human edits into machine-generated outputs. In our work, we leverage human-identified salient excerpts and use LLMs to generate contextual information, improving comprehension of social conversations. Additionally, we focus on much longer, multi-party conversations.

\section{Human-annotated Salient Excerpts Dataset (HSE)}
\label{sec:data}

To enable the salient excerpt-driven understanding of social group conversations, we create the \emph{Human-annotated Salient Excerpts Dataset (HSE)}\footnote{Accessible at \url{https://github.com/shresh02/bridging_context_gaps}}. This dataset is built on top of the Fora dataset~\cite{schroeder-etal-2024-fora}, which includes social group conversations organized by the non-profit 
\href{https://cortico.ai/}{Cortico} in partnership with various US organizations. These are small-group facilitated conversations that revolve around topics such as \emph{affordable housing, education, public health, and neighborhood safety}. Participants often share their perspectives and personal stories during these discussions. The dataset includes 262 conversations having an average of 66 minutes in length, where a conversation has a median of 6 participants and an average of 152 speaker turns. Figure ~\ref{fig:len_data} (Appendix) shows the distribution of conversation length within the dataset, with most conversations ranging from \emph{7,000 to 12,500 words}.

Humans were asked to highlight salient excerpts in each conversation. That led to 1,774 salient excerpts across 177 conversations in the Fora dataset. Each excerpt averages 111 words, with over 75\% tagged by topics such as civic engagement, educational inequality, and financial literacy.


These excerpts, while salient and representing important points in the conversation, often lack the necessary context for full comprehension. Our study focuses on effective contextualization of the excerpts. To be able to assess the models' performance in generating enhanced contexts, we annotated excerpts with social and contextual attributes from the conversation, such as speaker demographics, motivations, locations, and relevant events. We crowd-sourced annotators through the Prolific platform, providing them with 35 full social group conversations from the Fora dataset and 100 excerpts across these conversations. Annotators responded to specific questions listed in Appendix~\ref{appendix:hse_annotations}. After filtering out incomplete or inattentive responses based on task completion times, we retained approximately 90 excerpts, each annotated by 4-5 annotators. We conducted a secondary review of the annotations to ensure consistency, establishing a ground truth for evaluating the models' objective performance.

Given the need for developing socially aware models~\cite{yang2024call} and the limited availability of relevant datasets, the HSE dataset presents a valuable resource for advancing research in this area. Although the additional tags, such as topics, are not the primary focus of this paper, they can support other socially oriented tasks and benchmarks (e.g., topic detection). The excerpts can also be applied to summarization tasks, as demonstrated in a later section, to enhance overall comprehension of the full conversation.

\section{Problem Setup: Effectively Contextualizing Excerpts}


Understanding excerpts from long-form social group conversations could be non-trivial and time-consuming given their length. Our primary goal is to contextualize these excerpts to provide meaningful and useful information, which we term \textit{effective contextualization}. We formalize the problem as follows:

The data consists of long-form group conversations \( C = \{C_1, C_2, \dots, C_n\} \), where each \( C_j \) is a conversation among $K$ participants on some topic(s). For each \( C_j \), a set of salient excerpts (in our case human-highlighted) is available, $E_j = \{e_{j1}, e_{j2}, \cdots, e_{jm}\}$,  where $e_{jl}$ is a salient excerpt. For a given excerpt from a conversation, we aim to generate additional context so that the overall comprehensibility and understanding of the conversation are improved. 

Formally, $\mathbf{\tilde{e}}_{jl} = f(C_j, e_{jl})$ represents the \emph{context-enriched excerpt} for $e_{jl}$. Here, $f$ is an LLM that is used for extracting the relevant information and generating $\mathbf{\tilde{e}}_{jl}$. This process of generating the effective context for a given excerpt is referred to as \emph{effective contextualization}. 

The key questions here are: \emph{(1)} What additional information should be sought for effective contextualization in a social group conversation? \emph{(2)} How do we evaluate effective contextualization? 

\subsection{Factors in Effective Contextualization}
Our work is inspired by \cite{hovy2021importance} which uses linguistic frameworks such as systemic functional linguistics~\cite{halliday2013halliday} and the Cooperative Principle~\cite{grice1975logic}, proposing factors for effective communication in social conversational settings. We focus on the \textit{Speaker Characteristics} and \textit{Context} factors to improve the understanding of excerpts and specifically emphasize the following factors to improve the comprehensibility and usefulness of excerpts:

\begin{itemize} \setlength\itemsep{0em} \item \textbf{Speaker characteristics:} Demographic details (age, gender, ethnicity), educational, and occupational references, along with relevant personal or cultural experiences. \item \textbf{Speaker motivation:} Understanding the reasons behind a speaker's statements, including the context of the questions or comments they respond to. \item \textbf{Relevant locations or events:} Identifying key locations or events in the conversation to aid comprehension. \end{itemize}

\subsection{Evaluating Effective Contextualization}
Once an effectively contextualized excerpt $\mathbf{\tilde{e}}_{jl}$ has been generated, we evaluate the effectiveness of this contextualization by evaluating how well $\mathbf{\tilde{e}}_{jl}$ improves the comprehension and informativeness through the additional information. We also evaluate the relevance and consistency of the generated excerpt $\mathbf{\tilde{e}}_{jl}$ with respect to the original excerpt $e_{jl}$ and the conversation $C_j$. The overall evaluation is done through subjective evaluations involving humans as well as through objective faithfulness analyses. The subjective evaluations are done across dimensions such as understandability, readability, completeness, and cohesiveness. 

\subsection{Implicit and Explicit Effective Contextualization}
We utilize Large Language Models (LLMs) for the effective contextualization of conversation excerpts. For a given excerpt \( e \) from a conversation \( C \), we use an LLM to generate a \emph{Context-Enriched Excerpt (CEE)}, using the surrounding contexts within \( C \). We use two approaches:

\begin{itemize}
    \item \( CEE_i (e)\) represents the excerpt obtained through \textbf{implicit contextualization}, where zero-shot capabilities of the LLM~\cite{kojima2022large} are leveraged to generate context without explicit instructions on social attributes.
    \item \( CEE_e (e)\) represents the excerpt obtained through \textbf{explicit contextualization}, where the LLM is guided by \emph{In-Context Learning (ICL)}~\cite{brown2020language} and is explicitly instructed to incorporate specific social attributes from the conversation.
    
\end{itemize}

In both cases, the input consists of both the conversations and the excerpts. The LLM is tasked with gathering relevant information from the conversation, whether by focusing on explicit social attributes (in \( CEE_e \)) or by relying on inherent language patterns in the implicit case (in \( CEE_i \)). Prompts for both methods are provided in Appendix~\ref{appendix:prompts}.

More formally, $CEE_e (e_{jl}) = f_e(C_j, e_{jl}) $ and $CEE_i(e_{jl}) = f_i(C_j, e_{jl})$
represent the explicit and implicit context-enriched excerpts for $e_{jl}$ respectively.  \( f_e \) corresponds to the LLM's generation with an explicit focus on social attributes, and \( f_i \) corresponds to the zero-shot generation without explicit instructions.

\section{Experimental Setup}
\label{sec:experiment-setup}
We now describe the overall experimental setup and details of the evaluation approach. We investigate the effective contextualization of excerpts in the following three directions. 

\noindent \textbf{$\mathbf{CEE}_e \,\, \text{vs} \,\, \mathbf{CEE}_i$}: Our first empirical evaluation studies the difference in explicit vs implicit effective contextualization. More specifically, we conduct subjective evaluations to analyze and compare the excerpts and their context-enriched version ($CEE_e$ and $CEE_i$) across various dimensions such as understandability, readability, and cohesiveness (Table \ref{tab:avg_ratings_implicit_explicit}). The contextualization function $f$ here is \emph{GPT-4 Omni (GPT4-o)}~\cite{achiam2023gpt}. The details of the evaluation process are described in the subsequent section.

\noindent \textbf{LLMs for contextualization:} 
We next compare different LLMs for the task of effective contextualization. We compare 3 LLMs \emph{GPT-4 Omni (GPT4-o)}~\cite{achiam2023gpt}, \emph{Llama 3.1-70b}~\cite{dubey2024llama} and \emph{Claude Opus}~\cite{Anthropic_2024}. Llama 3.1-70b is an open-source LLM whereas the other two are not. Due to cost considerations, we limit our experimentation on Claude and Llama to only explicit contextualization\footnote{As we see later in Section \ref{sec:results}, this is the better performing effective contextualization approach. Given that we are restricted by cost considerations, we opted to limit the performance comparison of the 3 LLMs on this better performing case.} (as opposed to both implicit and explicit). This comparative analysis helps us evaluate the relative effectiveness of different state-of-the-art LLMs in generating meaningful and faithful context for conversation excerpts. Complete prompts and further details are provided in Appendix~\ref{appendix:prompts}.

\noindent \textbf{Clarification through extrinsic knowledge:}  
To further probe the efficacy of LLMs in generating context-enriched excerpts for social conversation, we investigate their ability to clarify uncommon terms and phrases which can further improve the generated excerpt. As part of the explicit contextualization method (\( CEE_e \)), we prompt the LLMs (Appendix~\ref{appendix:prompt_extrinsic_knowledge}) to clarify any specialized terms or phrases that might not be common knowledge by requiring the LLM to provide additional explanations or references to extrinsic knowledge to enhance the comprehensibility of the context. We analyze how well the LLMs can identify such terms and provide useful clarifications, particularly when using models like GPT-4 Omni's web search capability. This capability was expected to enrich the context by offering definitions that could otherwise be unclear to the reader.

\subsection{Evaluation}
\label{sec:evaluation}
We conduct an evaluation of the enriched context for 90 annotated excerpts from the HSE dataset (Section~\ref{sec:data}) through a combination of subjective human assessments and objective faithfulness measures. We assessed the quality of the enriched excerpts produced by various models, focusing on dimensions such as \textit{faithfulness, text quality}, and \textit{speaker perception}.\newline

\noindent \textbf{Subjective human evaluation:}  
We recruited 75 human evaluators from Prolific with a 99-100\% approval rating to evaluate the enriched contexts. Participants assessed the excerpts on textual quality (\textit{understandability, readability, redundancy, completeness, cohesiveness}) and speaker perception (\textit{agreement with the speaker's point of view, perception of the speaker as honest or trustworthy, respect for the speaker, empathy for the speaker, the ability to see the speaker’s point of view (POV)}). This evaluation determined how well the enriched contexts convey the speaker’s original message and their social perception.

Each excerpt was rated by at least three evaluators. Quantitative data was collected using likert scale ratings [1-5], and qualitative insights were gathered through open-ended responses. Details of the survey and prompts are available in Appendix~\ref{appendix:survey_scnshots}.

\noindent \textbf{Faithfulness and objective metrics:}  
To measure the consistency of the enriched context with the source conversation, we assess the faithfulness~\cite{li2022faithfulness} of the information extracted from the conversations and compare it with the annotated social attributes from the HSE dataset. We categorize the extracted attributes into short-response factors and long-response factors, based on the type and depth of the information being conveyed:

\begin{itemize}
    \item \textbf{Short-response factors} are brief, factual details like the speaker’s name, gender, age, occupation, race, and economic status. These require minimal elaboration and can typically be conveyed in a few words.
    \item \textbf{Long-response factors} involve more nuanced and context-rich information, such as the speaker’s background, personal experiences, significant events, or locations. These require detailed explanations to capture complexity. For instance, locations could be specified at multiple levels (e.g., city, state, country), adding depth to the context, which makes them long-response factors.
\end{itemize}

We calculate the F1-score to balance precision and recall, where precision was measured by the cosine similarity between the SBERT embeddings~\cite{reimers2019sentence} of extracted attributes from the context and the annotations in the HSE dataset, and recall was determined based on whether essential information was missed. The extraction of attributes was done using GPT-4o and details are provided in Appendix~\ref{appendix:faithful_extraction}
\section{Results and Discussion}
\label{sec:results}

\begin{table}[ht]
    \centering
    \resizebox{\columnwidth}{!}{
    \begin{tabular}{p{6cm}| c c c}

     \textbf{Dimension}& e & $CEE_{i}$ & $CEE_{e}$\\
    
    \hline
    Understandability & 3.43 & 4.11* & \textbf{4.13*} \\
    Readability & 3.40 & 3.90* & \textbf{4.10*$\dagger$} \\
    Low redundancy & 3.24 & 3.53* & \textbf{3.75*$\dagger$} \\ 
    Completeness & 3.07 & \textbf{3.96*} & 3.83*$\dagger$ \\ 
    Cohesiveness & 3.23 & 3.93* & \textbf{4.03*}  \\
    \hline
    Agreement with the speaker's p.o.v. & 3.57 & \textbf{4.01*} & 3.99* \\ 
    Viewing speaker as honest/trustworthy & 3.55 & 3.91* & \textbf{3.98*}\\
    Respect for speaker & 3.58 & 3.89* & \textbf{3.97*} \\
    Empathy for speaker & 3.50 & 3.59 & \textbf{3.66*} \\
    Ability to see the speaker’s p.o.v. & 3.77 & \textbf{4.08*} & 4.06* \\
    \hline
    \end{tabular}
    }
    \caption{Average ratings from the human evaluation of the original excerpt \textit{e}, and context enriched excerpts \textit{$CEE_i$} (implicit) and \textit{$CEE_e$} (explicit) generated using GPT-4o where * denotes a statistically significant difference for the ratings from the original excerpt and $\dagger$ shows a statistically significant difference between the ratings for \textit{$CEE_i$} and \textit{$CEE_e$} for Welch t-test for $p < 0.05$. The horizontal line divides dimensions grouped into two broader categories: textual quality (upper half) and speaker perception (lower half).} 
    \label{tab:avg_ratings_implicit_explicit}
\end{table}

\begin{table}[ht]
    \centering
    \resizebox{\columnwidth}{!}{
    \begin{tabular}{p{6cm}| c c c }
    
    \textbf{Dimension} & GPT & Claude & Llama\\
    
    \hline
    Understandability & 3.70 & \textbf{3.81} & 3.58$\dagger$ \\
    Readability & 3.86 & \textbf{3.90*} & 3.71*$\dagger$ \\
    Low redundancy & 3.66 & \textbf{3.72*} & 3.51*$\dagger$ \\ 
    Completeness & 3.67 & \textbf{3.72} & 3.58\\ 
    Cohesiveness & 3.66 & \textbf{3.81*} & 3.47*$\dagger$ \\
    \hline
    Agreement with the speaker's p.o.v. & 4.07 & \textbf{4.10} & 4.00 \\ 
    Viewing speaker as honest/trustworthy & 4.07 & \textbf{4.17} & 4.02$\dagger$ \\
    Respect for speaker & 4.19 & \textbf{4.21} & 4.18 \\
    Empathy for speaker & 4.19 & \textbf{4.22} & 4.18$\dagger$ \\
    Ability to see the speaker’s p.o.v. & 4.20 & \textbf{4.29} & 4.11$\dagger$\\
    \hline
    \end{tabular}
    }
    \caption{Average ratings from the human evaluation of explicitly enriched contexts generated by GPT, Claude, and Llama where * denotes a statistically significant difference between the ratings for Llama and Claude against GPT and $\dagger$ denotes a statistically significant difference between the mean ratings for Llama and Claude for the Welch t-test for $p < 0.1$. The horizontal line divides dimensions grouped into two broader categories: textual quality (upper half) and speaker perception (lower half).} 
    \label{tab:avg_ratings_llm}
\end{table}

\subsection{Textual quality and Speaker Perception}

\textbf{$\mathbf{CEE}_e \,\, \text{vs} \,\, \mathbf{CEE}_i$}: A total of 90 annotated excerpts from the HSE dataset were used for this analysis, and each excerpt was rated by an average of 4-5 crowdsourced evaluators, resulting in 433 total responses. On average, each excerpt contained 128 words. The implicit contextualization (\(CEE_i\)) contexts have an average length of 267 words, while the enriched (\(CEE_e\)) contexts average around 55 words. This difference arises from the approaches: \(CEE_i\) includes all relevant details from the full conversation, whereas \(CEE_e\) focuses solely on extracting key social factors.

In Table~\ref{tab:avg_ratings_implicit_explicit}, we observe that both methods of contextualization received significantly higher ratings than the original salient excerpts alone (p < 0.05, Welch t-test). Moreover, the explicit contextualization (\(CEE_e\)) method generally outperformed implicit contextualization (\(CEE_i\)) on most dimensions. The ratings for \(CEE_e\) were particularly high in comprehensibility, readability, low redundancy, and cohesiveness as well as in cultivating respect and empathy for the speaker. This was attributed to the inclusion of key social factors that improved the understanding of the excerpts. In contrast, participants found \(CEE_i\) more complete, as it included more details from the conversation. Notably, explicit contextualization results in shorter enriched contexts while improving performance across several dimensions.

\noindent \textbf{LLMs for contextualization:}  We compare three LLMs, GPT-4 Omni (GPT-4o), Claude Opus, and Llama 3.1-70b, on their ability to generate enriched contexts using \(CEE_e\). We recruited 30 evaluators to rate the contexts generated by each model. Each excerpt was rated by an average of three evaluators. 
Claude emerged as the most preferred model, achieving the highest ratings across nearly all dimensions, including understandability, readability, and speaker perception (Table \ref{tab:avg_ratings_llm}). 

The enriched contexts varied in length among the models. GPT-4o produced contexts averaging 296 words, while Claude Opus generated slightly shorter contexts, averaging 264 words. In contrast, Llama 3.1-70b produced notably longer contexts, averaging 538 words. 
The differences between Llama 3.1-70b and the other models were statistically significant (p<0.05), particularly in areas like understandability, readability, redundancy, empathy, and the ability to see the speaker’s POV.

\subsection{Qualitative Analysis of Responses}
 Beyond the quantitative evaluation, the comments from the evaluators provided nuanced insights into how the inclusion of social attributes and background information influenced their perceptions. We provide some qualitative insights through these comments. 
 

Evaluators emphasized the value of background information such as the speaker's name and other contextual cues, which helped them connect more deeply with the speaker while preferring conciseness. This enhanced trustworthiness, empathy, and the overall utility of the contexts. One participant stated, \textit{“Giving background information about the speaker gave it a personal touch and helped me empathize with the reader more.”} Another participant echoed this sentiment, saying that the \(CEE_e\) context \textit{“succinctly introduces [speaker name] as a high school student from Gardiner, Maine, sharing her thoughts on essential career skills such as collaboration, writing, and speaking.”}

The preference for \(CEE_e\) was frequently attributed to its ability to offer a broader and more socially rich context without excessive detail. As noted earlier, \(CEE_e\) produces considerably shorter enriched contexts and is yet preferred on factors such as readability and cohesiveness.  One participant commented, \textit{“I preferred the \(CEE_e\) context because it offers a broader context for the excerpt, explaining the community's overall goals and how the speaker's concerns fit into those goals. This context makes the excerpt more meaningful and easier to understand.”} This underscores how \(CEE_e\) successfully situates the excerpt within the larger conversation while maintaining conciseness, thus enhancing the overall understandability.

We also observe that disagreement with the speaker's POV could coexist with high ratings in empathy and the ability to understand the speaker's perspective. In several cases, evaluators rated their agreement with the speaker's perspective as low (1-2), while still giving high ratings (3-5) for empathy and related perception questions. The \(CEE_e\) context had slightly more such cases, with 15 instances of high empathy and the ability to see the speaker's POV despite disagreement, compared to 11 cases of high empathy in the \(CEE_i\) context.

\begin{figure}[h]
    \centering
    \includegraphics[width=0.45\textwidth]{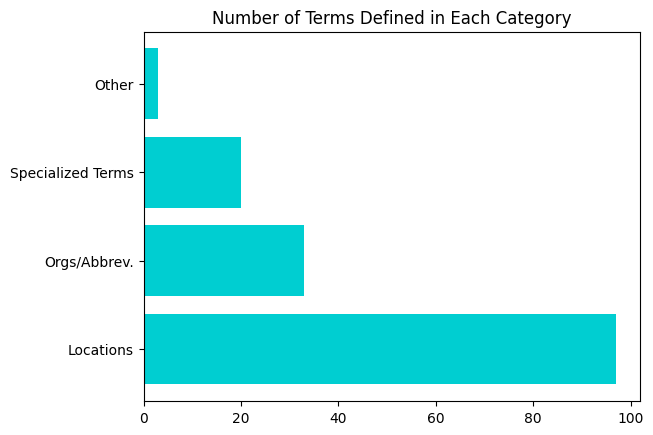}
     \caption{{Number of defined terms in each category}}
     \label{fig:terms_cat_plot}
     \vspace{-0.15in}
\end{figure}

\subsection{Clarification through Extrinsic Knowledge }

As part of the explicit contextualization (\(CEE_e\)) process, we prompted the LLMs to clarify uncommon or specialized terms to enhance comprehensibility by incorporating extrinsic knowledge not directly mentioned in the conversation. This included providing additional context about locations (e.g., Cornwallis, Durham), explaining abbreviations (e.g., a school district acronym or a church's name), or elaborating on specialized terms not defined in the conversation (e.g., "green roofs," "heat island"). Figure~\ref{fig:terms_cat_plot} illustrates the distribution of these categories. The "Other" category refers to terms that did not fall into the three prevalent categories. Terms in this category were largely names of other participants in the conversation. Of the 433 responses collected during the human evaluations, 128 responses included specifically flagged terms that required elaboration.  The \(CEE_e\) process successfully addressed only \textbf{55} of these instances, highlighting that LLMs still struggle to consistently identify and explain uncommon terms critical for tailoring content to the reader. However, when the models correctly defined terms using extrinsic knowledge, the evaluators found the elaboration helpful, particularly for specialized terms that could otherwise cause confusion, which helped in improving their understanding of the excerpt.


\subsection{Faithfulness Analysis}

It is important that the enriched contexts remain consistent with the original conversation. We evaluate the consistency of the enriched contexts with the original conversation through an objective faithfulness analysis, focusing on both short-response and long-response factors (Table \ref{tab:gpt_factuality}). Both \(CEE_e\) and \(CEE_i\) performed well in extracting short-response information, such as the speaker's name, gender, and race, demonstrating high precision. The explicit contextualization (\(CEE_e\)) exhibited slightly higher recall rates, suggesting it was more adept at capturing a broader range of relevant details, aligning with its design to identify social attributes efficiently. However, the LLMs struggled with long-response factors, such as speaker background, shared personal experiences, and location context, likely due to the high variability and open-ended nature of these elements suggesting areas for future improvement. Despite this challenge, \(CEE_e\) showed a stronger ability to extract fine-grained information, as it focused on highlighting key attributes within the conversation. 
Comparisons of GPT-4o, Claude Opus, and Llama 3.1-70b for faithfulness are detailed in Appendix~\ref{appendix:faithfulness_models}. Claude Opus generated the most faithful responses.

\begin{table}
    \centering
    \resizebox{0.5\textwidth}{!}{  
    \begin{tabular}{p{6cm} c c }  
    \hline
    \textbf{Factor} & $CEE_{i}$ & $CEE_{e}$ \\
    \hline
    Name of speaker & 0.87 & \textbf{0.93} \\
    Gender & {0.96}  & \textbf{0.98} \\
    Race & \textbf{0.91} & \textbf{0.91} \\
    Education & 0.85  & \textbf{0.90} \\ 
    Occupation & {0.71} & \textbf{0.86} \\
    Age & 0.88   & \textbf{0.89} \\ 
    Economic status & 0.87 & \textbf{0.88}  \\
    \hline
    Additional info on speaker background & 0.43  &\textbf{0.69} \\ 
    Personal experiences & 0.41 & \textbf{0.48}  \\ 
    Question/response that prompted speaker & \textbf{0.74} & 0.52 \\
    Significant events/locations mentioned & \textbf{0.77} & 0.68  \\
    Location of conversation & \textbf{0.51} & 0.34  \\
    \hline
    \end{tabular}
    }
    \caption{{F1 score of $CEE_{i}$ and $CEE_{e}$ enriched contexts for Faithfulness Analysis}. Factors are divided into two broader categories: short-response factors (upper half) and long-response factors (lower half).}
    \label{tab:gpt_factuality}
    \vspace{-0.1in}
\end{table}

\section{Context-Enriched Excerpts for Enhanced Comprehension of Conversation}
\label{sec:context_comprehension}
Intuitively, highlighted salient excerpts could be useful in highlighting key information or providing some form of a summary of the full conversation. To evaluate whether context-enriched excerpts lead to better summaries, we compare the effectiveness of these excerpts against direct summarization of the full conversations. Specifically, we generate and compare three types of summaries (using GPT-4) with a word limit of 200 for each:

\begin{itemize}
    \item \textbf{Full Conversation Summary}: Generated by providing the LLM with the entire conversation.
    \item \textbf{Excerpt-Only Summary}: Generated using only the salient excerpts identified for the conversation.
    \item \textbf{Context-Enriched Excerpt Summary}: Generated using all excerpts after they have been effectively contextualized.
\end{itemize}

This comparison aims to determine whether enriching excerpts with context can effectively capture the key points of the conversation and enhance overall comprehension within the set word limit. 

A total of 50 conversations were analyzed, with 3-4 evaluators rating each summary, resulting in 156 responses from 34 evaluators. Table \ref{tab:sum_comparison} shows the average ratings on various dimensions. The Context-Enriched Excerpt Summary received the highest ratings in coherence, informativeness, and level of detail, outperforming both the Full Conversation Summary and the Excerpt-Only Summary. While the Excerpt-Only Summary scored slightly higher in readability—likely due to its straightforward merging of salient excerpts—it was rated lower in informativeness and detail, as it lacked the additional context from the conversation. Participants noted that the Context-Enriched Excerpt Summary effectively captured the main points while providing essential background information, thereby enhancing overall comprehension.

A key strength cited in the qualitative responses for this approach was that the additional details including names and background information of speakers induced empathy and connection. As one participant noted, it \textit{"gives light to names of those in school and their educators... It gives you a reason to get behind these people because you feel close to them and can see their struggle."} . This underscores the importance of effectively contextualizing salient excerpts with social aspects to convey the essential elements of the conversation, offering a focused yet comprehensive understanding.



\begin{table}[t]
    \centering
    \setlength\tabcolsep{3pt}
    \resizebox{\linewidth}{!}{
    \begin{tabular}{p{3.5cm} c c c}

    \textbf{Dimension} & \textbf{Full Conv.} & \textbf{Excerpt-Only} & \textbf{CEE} \\
 
    \hline
    Readability & 4.21 & \textbf{4.25} & 4.11 \\ 
    Coherence \& structure & 4.04 & 3.98 & \textbf{4.19} \\ 
    Informativeness & 4.16 & 3.86 & \textbf{4.26} \\
    Detail level & 4.06 & 3.66 & \textbf{4.16} \\ 
    \hline
    \end{tabular}
    }
    \caption{Average ratings for different summarization approaches of the conversation.}
    \label{tab:sum_comparison}
    \vspace{-0.2in}
\end{table}

\section{Conclusion}
\label{sec:conclusion}
Long-form social group conversations often convey rich information but are challenging to process due to their length. Human-highlighted salient excerpts provide anchor points for understanding but can lack sufficient context, potentially leading to misunderstandings, particularly given the informal and personal nature of these conversations. To address this, we introduced a method using LLMs to augment salient excerpts with effective contextualization. This approach significantly improved comprehension, readability, and empathy by enriching excerpts with meaningful context. While LLMs show promise, challenges remain in capturing speaker background, personal experiences, contextualizing locations, and addressing unique terminology. We also demonstrated the utility of contextualized excerpts for summarizing lengthy conversations, with subjective evaluations highlighting their improved effectiveness. To support further research, we release the Human-annotated Salient Excerpts (HSE) dataset as a resource for advancing understanding and contextualization in social conversations.

\section{Limitations}

\textbf{Consistency of generated context}: LLMs can produce different results with each iteration. To ensure consistency, we set the model temperature to 0 for deterministic outputs and collected a large number of responses to accurately reflect the representative average rating from the population.

 \textbf{Faithfulness evaluation}: F1 scores are lower for open-ended questions due to their subjective nature, which complicates standardization. For instance, responses about a speaker's background varied among annotators, highlighting the challenge of standardizing such tasks. Further research is needed to develop better methods and metrics for measuring faithfulness, as noted by \cite{risch2021semantic} and \cite{xu2024reasoning}. Our final method for assessing faithfulness was chosen as other methods, such as Named Entity Recognition and entailment approaches \cite{goyal2021annotating, manakul2023selfcheckgpt}, were less suitable for our specific use case.


\textbf{Number of excerpts}: Due to cost constraints, we analyzed only 90 excerpts. Each required multiple ground truth annotators to get accurate ground truth data, making the process expensive.



\section{Ethics Statement}

\textbf{Code of Conduct} The informed consent of all survey participants and annotators was obtained before participation. This study received approval from the Institutional Review Board (IRB) at our university, and participants were compensated at a rate above the minimum wage for the state of Massachusetts.

\textbf{Privacy of Models}: We are aware of the privacy concerns associated with releasing information to LLM models for analysis. We used GPT and Claude via APIs, and both OpenAI and Anthropic have stated that data sent to the API would not be used for training \cite{openaiuse,claudedatause}. As a note, the Fora dataset \cite{schroeder-etal-2024-fora} upon which we build our HSE dataset involves conversation data that has received the proper consent to be publicly released and underwent anonymization to further preserve the privacy of the conversation participants.

\section*{Acknowledgements}
All experimentation and data processing were carried out by the authors at MIT. We thank everyone at Cortico for their collaboration in curating the HSE dataset. We also thank Hang Jiang and Daniel Kessler for their feedback on the evaluation design.

\bibliography{anthology}

\onecolumn
\appendix

\section{Faithfulness evaluation of LLM enriched contexts}
\label{appendix:faithfulness_models}

When comparing the different LLMs using the \(CEE_e\) approach, we found that GPT-4o and Claude Opus demonstrated slightly better precision and recall for short-response factors than Llama 3.1-70b, as shown in Table \ref{tab:llm_factuality}. in the Appendix. However, Llama tended to capture more details related to personal experiences and the preceding conversation topics, possibly due to its generally longer responses. This verbosity, while occasionally helpful, sometimes led to a decrease in overall cohesiveness and readability, reflecting the need for a careful balance in context generation.

\begin{table}
    \centering
    \resizebox{0.5\textwidth}{!}{  
    \begin{tabular}{p{6cm} c c c }  
    \hline
    \textbf{Factor} & $GPT$ & $Claude$ & $Llama$ \\
    \hline
    Name of Speaker & 0.94 & 0.95 & \textbf{0.97} \\
    Gender & \textbf{1.00}  & \textbf{1.00} & \textbf{1.00} \\
    Race & \textbf{0.98} & \textbf{0.98} & \textbf{0.98} \\
    Education & \textbf{0.97}  & \textbf{0.97} & 0.96 \\ 
    Occupation & \textbf{0.90} & 0.88 & 0.89 \\
    Age & {0.97}   & \textbf{0.98}  & 0.83\\ 
    Economic Status & 0.73 & 0.70 & \textbf{0.85}  \\
    \hline
    Additional info on speaker background & 0.80  &\textbf{0.81} & 0.78 \\ 
    Personal Experiences & \textbf{0.78} & 0.77 & 0.76 \\ 
    Question/Response that prompted speaker & 0.80 & 0.83 & 0.82 \\
    Significant Events/Locations mentioned & 0.88 & \textbf{0.90} & 0.86  \\
    Location of Conversation & \textbf{0.77} & 0.71 & \textbf{0.77}  \\
    \hline
    \end{tabular}
    }
    \caption{\textbf{F1 score of GPT-4o, Claude Opus, and Llama 3.1 enriched contexts for Faithfulness Analysis}. The horizontal line divides the factors into two broader categories: direct, short-response factors (upper half) and open, long-response factors (lower half)}
    \label{tab:llm_factuality}
    \vspace{-0.1in}
\end{table}

\section{Examples of Explicit Effective Contextualization of Excerpts}
\label{appendix:examples_effective_contextualization}

\begin{longtable}{|p{3cm}|p{3cm}|p{3cm}|p{3cm}|}
\caption{Examples of Effective Contextualization} \label{appendix:examples_effective_contextualization} \\
\hline
\textbf{Original Excerpt} & \textbf{Generated Context} & \textbf{Annotator Comments} & \textbf{Social Attributes Present} \\ 
\hline
\endfirsthead

\hline
\textbf{Original Excerpt} & \textbf{Generated Context} & \textbf{Annotator Comments} & \textbf{Social Attributes Present} \\ 
\hline
\endhead

\hline
\endfoot

\hline
\endlastfoot

I just have one quick thing and it's the concern for our teachers. I have a lot of concern for teachers leaving, good teachers, leaving the industry because for financial reasons ... And I have teachers in my family, I have teachers that are clients, I have teachers that are friends, and they're exhausted and we're losing the good ones ... 
& \textit{\textbf{[Speaker name]}}, a parent from Kennebec County, expressed her deep concern about the exodus of quality teachers from the education sector ... where participants shared their personal experiences and visions for the future of education in their community. 
& I found the second version of the excerpt and accompanying information to be the most informative and clear. It provided relevant details about \textit{\textbf{[speaker name]'s}} concerns regarding teachers leaving due to financial reasons and inadequate retirement programs, without being redundant or vague ... offering a comprehensive understanding of the excerpt's context and significance.
& \begin{itemize}
    \item Additional information on speaker's background
    \item Speaker's Name
    \item Personal Experience
\end{itemize}
\\ \hline

... And then the greater question then becomes, what happens to community when the folks who are from this place are no longer of this place? It can go down, that's the avenue, and there are all these new really big houses, which is great ... They don't even have a backyard. It's just the house. There's no front yard, just a house. And so the purpose is in order to be as isolating and as isolated as humanly possible...
& \textbf{[Speaker name]}, a pastor at \textit{\textbf{[church name]}} in \textit{\textbf{Durham}}, has been receiving increasing calls for rental assistance since the pandemic, highlighting the severe housing crisis in the area; \textit{\textbf{he moved to Durham in 2018 to pastor and has witnessed the city's demographic and socioeconomic shifts, which have exacerbated housing instability and poverty.}}
& I chose to focus on understanding the broader context and details of the housing and community issues discussed by \textit{\textbf{[speaker name]}} because it provides a deeper insight into the underlying causes, impacts, and potential solutions to the problems he highlights, enriching the interpretation of the excerpt. This additional information helps contextualize his concerns and vision within the larger socioeconomic landscape of \textit{\textbf{Durham}}.
& \begin{itemize}
    \item Personal experience
    \item Location
    \item Speaker's name
\end{itemize}
\\ \hline

I want to help my community by making it so everyone in my community is able to be in affordable housing, whether that's giving fundraisers as well, I'm going to be an electrician when I grow up so I can help them give discounts and I think that's just going to help everyone continue on with their lives better.
& \textit{\textbf{[Speaker name]}}, a participant \textit{\textbf{from Kennebec County, Pittston}}, shared his aspiration to support his community by ensuring access to affordable housing through fundraisers and offering discounts as an \textit{\textbf{electrician}}; this was in response to [participant name]'s question about \textit{\textbf{how they want to contribute to their community and how their career fits into that}}.
& This version provided \textit{\textbf{context about the conversation}}, \textit{\textbf{[speaker name]'s}} thoughts and motivations, and clarified how his career as an \textit{\textbf{electrician}} will help with affordable housing, making it easy to understand his perspective and goals without needing additional information.
& \begin{itemize}
    \item Speaker's name
    \item Occupation
    \item Question / Response that motivated the speaker
\end{itemize}
\\ \hline

... and there may be special items needed to bring that about. I think that's something that we can work on. Just like, for instance, what's done with museums, sometimes people with vision can go and see and hear, and do everything. But for us, they sometimes have some kind of technology that makes it accessible for us.
& The speaker, \textit{\textbf{[speaker name]}}, is \textit{\textbf{visually impaired}} and is advocating for the inclusion of facilities and activities that cater to the needs of visually impaired or deaf individuals; she emphasizes the importance of accessibility technology in public spaces, similar to what is available in museums.
& The added context about the committee's goals and \textit{\textbf{[speaker name]’s}} role provided a clear understanding of the issues being addressed and the significance of her request, helping to see the relevance of her comments in the larger discussion about accessibility.
& \begin{itemize}
    \item Speaker's name
    \item Additional information on speaker's background
    \item Personal experiences
\end{itemize}
\\ \hline

\end{longtable}

\section{Datasets Statistics}

\begin{figure}[H]
\centering
\begin{subfigure}{0.4\linewidth}
  \centering
  \includegraphics[width=\linewidth]{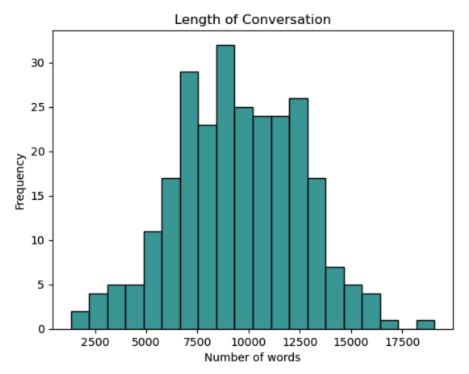}
  \label{fig:data_1}
\end{subfigure}\hfill
\begin{subfigure}{0.4\linewidth}
  \centering
  \includegraphics[width=\linewidth]{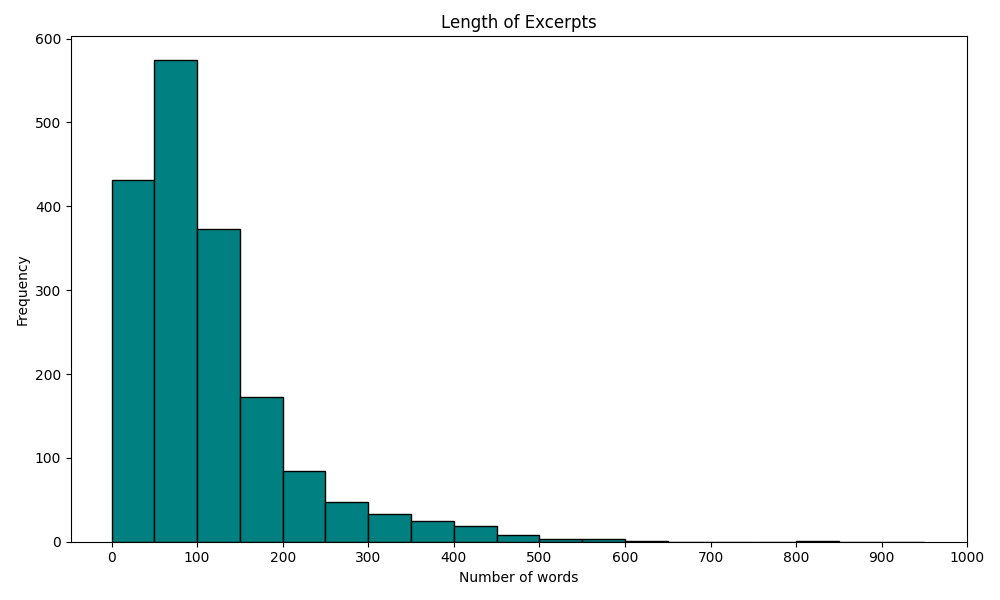}
  \label{fig:data_2}
\end{subfigure}
\caption{a) Length of conversation (in words) in Fora
dataset and b) Length of excerpts (in words) in human-annotated excerpts dataset}
\label{fig:len_data}
\end{figure}

\section{Prompts}
\label{appendix:prompts}
    
\subsection{Prompt for $CEE_i$ Context Generation}
\textbf{System Prompt}
\begin{lstlisting}
Summarize and contextualize an excerpt using the full conversation transcript for enhanced understanding and significance. 
Your goal is to build a foundation for the excerpt that fosters reader comprehension. 
Additional information should only be retrieved from the full conversation transcript. 
Do not make assumptions if the information is not explicitly mentioned in the conversation.
\end{lstlisting}

\textbf{User Prompt}
\begin{lstlisting}
Reflect on the main message or idea conveyed in the excerpt. Consider the full conversation and the broader context in which the excerpt occurred. 
Add additional information to better understand the significance of this excerpt.
The transcript of the full conversation is:  
{conversation} 
The excerpt is:  
{text}
\end{lstlisting}

\subsection{Prompt for $CEE_e$ Context Generation}

\textbf{System Prompt}
\begin{lstlisting}
Summarize and contextualize an excerpt using the full conversation transcript for enhanced understanding and significance. 
Your goal is to build a foundation for the excerpt that fosters reader comprehension. 
Additional information should only be retrieved from the full conversation transcript. 
Do not make assumptions if the information is not explicitly mentioned in the conversation.
You should also add justification of the context added in the summary of the excerpt. 
The objective is to enhance clarity by supplying both context and its rationale, aiding readers in fully comprehending the discussion's meaning and importance.
\end{lstlisting}

\textbf{User Prompt}
\begin{lstlisting}
Reflect on the main message or idea conveyed in the  excerpt. Consider the full conversation and the broader context in which the excerpt occurred. 
Output the following in three separate lines without the headers. 
It is important that there is ONLY ONE LINE for each of the following points SEPARATED BY A NEW LINE:
1.Add additional information from the conversation which would help better understand the excerpt and its significance in the broader context of the conversation. If the below is outputted in multiple lines, join them by a ;
Some of the items that could be added are:
- Background details of the speaker such as: name, demographic(such as age, gender), cultural, ethnic (including race), educational, occupational and economic references made by or about the speaker in the conversation.\
Do not make assumptions about the speaker if not mentioned explicitly by the speaker or referred to directly about the speaker.
- Motivation or intent of the speaker as to why they said what they said. At which point in the conversation was the excerpt brought up. To what question or statement was the excerpt a response.
- Relevant locations or events mentioned in the conversation that could help ground the excerpt better and make it easier to understand such as the location of the conversation.
- Also, include background information about where the speaker is currently staying or where they grew up, if it would help in understanding the excerpt better.
- Share any specific personal experiences mentioned by the speaker that is related to the excerpt in the conversation.
2. Include exact sentences from the conversation that support all of the additional information added. Cite the statements, speakers, and line numbers that support these additional anecdotes. Join them by a ;
3. Reasoning of added information: Explain why the added contextual information is crucial for a deeper understanding of the excerpt and its significance. Put this in one paragraph. 
The transcript of the full conversation is:  
{conversation) 
The excerpt is:  
{text}
\end{lstlisting}

\subsection{Prompt for Terms Clarification using Extrinsic Knowledge Retrieval}
\label{appendix:prompt_extrinsic_knowledge}
\textbf{System Prompt}
\begin{lstlisting}
You are a knowledgeable assistant that helps identify and define terms and locations that might not be common knowledge. 
For each of these, look them up on the web and provide a concise definition or explanation (1-2 lines) that clarifies their meaning or significance within the context of the text.
\end{lstlisting}

\textbf{User Prompt}
\begin{lstlisting}
I have a piece of text that may contain some terms and locations which might not be common knowledge. 
Please identify those terms and locations that are less likely to be familiar to the general public. 
For each of these, look them up on the web and provide a concise definition or explanation (1-2 lines) that clarifies their meaning or significance.
Multiple terms should be joined by a ; and the output should be in one line.


Example:
Text: 
The speaker expressing concern about safety is Peter, a third-grade teacher from Cumberland County. The conversation is part of a larger discussion about the state of education, specifically in Maine, and even more specifically in MSAD 11.

Output:
Term: The term "MSAD 11" refers to Maine School Administrative District 11, which is a school district in Maine that includes Gardiner, Pittston, Randolph, and West Gardiner.
Term: The term "Cumberland County" refers to a county in the state of Maine, which is located in the United States.

Here is the text that needs to be considered:

{text}
\end{lstlisting}

\subsection{Prompts for Faithfulness Factor Extraction}
\label{appendix:faithful_extraction}
\textbf{System Prompt}
\begin{lstlisting}
You are an expert tasked with answering questions about a context paragraph. 
    Follow these guidelines:
    1.  Do not make assumptions beyond the provided text.
    2.  Do not quote the question. 
\end{lstlisting}

\textbf{User Prompt}
\begin{lstlisting}
From the context below, answer the following questions:
Context: {context}

If any of the responses are not mentioned, please indicate that it is not mentioned. Do not quote the question in the answer.

Questions:

4.What is the name of the speaker? If the name is not directly mentioned but the number is, please specify the number only. If both are mentioned, mention the name only. If not mentioned, please answer "Not mentioned" exactly.
5.What is the gender of the speaker? If not mentioned, please answer "Not explicitly mentioned by the speaker" exactly.
6.What is the speaker's gender? If not, please answer "Cannot be inferred from the conversation" exactly.
7.What is the race mentioned by the speaker? If it is not mentioned, please answer "Not mentioned" exactly. 
8.What is the speaker's educational background? If it is not mentioned, please answer "Not mentioned" exactly.
9.What is the speaker's occupation? If it is not mentioned, please answer "Not mentioned" exactly.
10.What is the speaker's age? If it is not mentioned, please answer "Not mentioned" exactly.
11.Indicate the speaker's economic status, such as low-income, middle-income, high-income, unemployed, retired, student, or self-employed, without making assumptions beyond the provided transcript. If it is not mentioned, please answer "Not mentioned" exactly.
12.Is there any additional information about the speaker's background, such as cultural details, ethnic background, or where they grew up? If yes, please specify. 
14.Share any specific personal experiences mentioned by the speaker that are related, if applicable. If none, please answer "None".
15.What is the question or response from another person that prompted the speaker's statement? If the exact question or response is not given but the current conversation topic is, give the topic. Otherwise, please answer "Not mentioned".
16.Are there any significant events or locations that would provide additional context for understanding? If it is not mentioned, please answer "No".
17.Where did the group conversation take place? Please specify the county, city, state, or country if mentioned. If it took place over Zoom, you can specify that it is on Zoom, but also mention the city, state, or county. If not mentioned, please answer "Not mentioned".
\end{lstlisting}

\section{Demographics of Participants}
\label{appendix:demographics}

\subsection{Demographics of Participants of Effective Contextualized Excerpts - Human Evaluation}
\label{appendix:demographics_gpt}
\begin{figure}[H]
\centering
 \begin{subfigure}{0.4\textwidth}
     \centering
     \includegraphics[width=\textwidth, height=6cm]{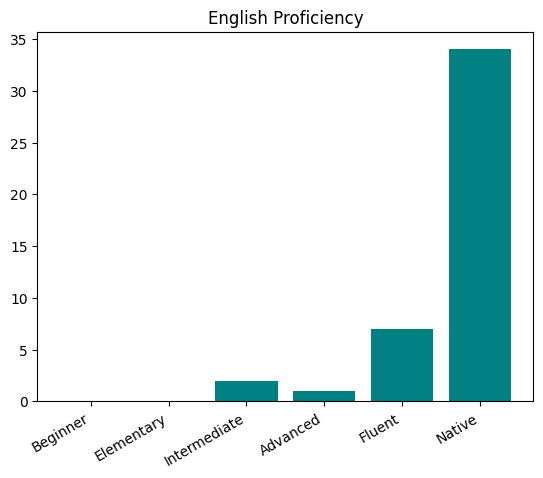}
     \label{fig: English Proficiency Bas v ICL}
 \end{subfigure}
 \hfill
 \begin{subfigure}{0.4\textwidth}
     \centering
     \includegraphics[width=\textwidth, height=6cm]{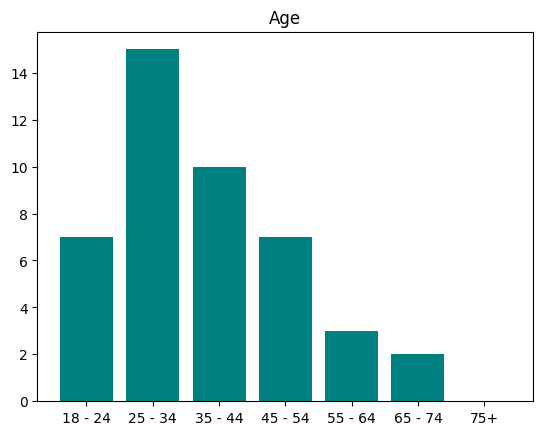}
     \label{Age Bas v ICL}
 \end{subfigure}
 
 \begin{subfigure}{0.4\textwidth}
     \centering
     \includegraphics[width=\textwidth, height=6cm]{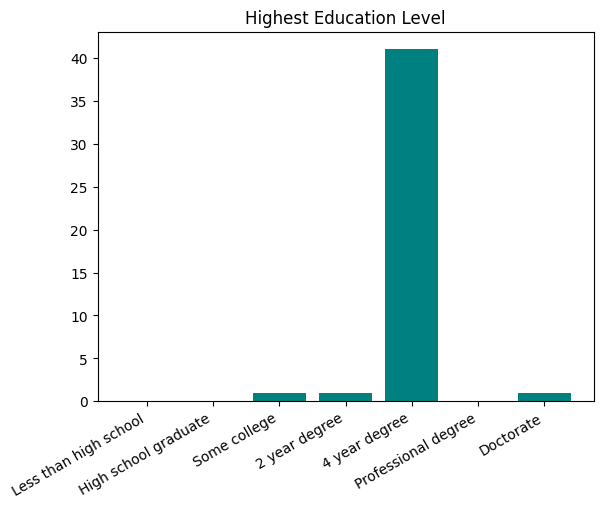}
     \label{Edu Bas v ICL}
 \end{subfigure}
 \hfill
 \begin{subfigure}{0.4\textwidth}
     \centering
     \includegraphics[width=\textwidth, height=6cm]{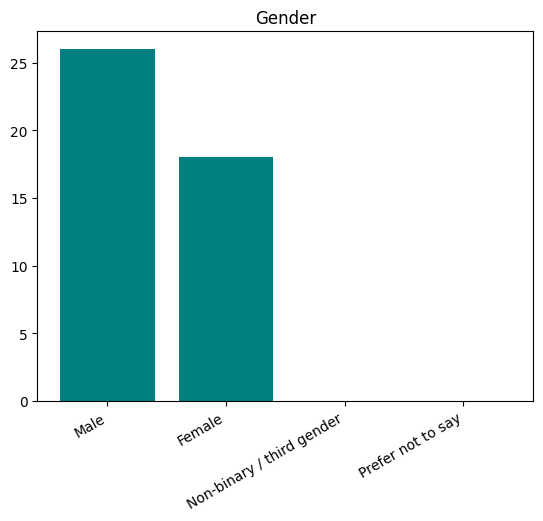}
     \label{Gender Bas v ICL}
 \end{subfigure}
    \caption{GPT Prompts Comparison Survey Demographics}
    \label{Bas v ICL Survey Demographics}
\end{figure}

\newpage
\subsection{Demographics of Participants of LLM Contexts Comparison}
\label{appendix:demographics_llms}
\begin{figure}[h!]
     \centering
     \begin{subfigure}{0.4\textwidth}
         \centering
         \includegraphics[width=\textwidth, height=6cm]{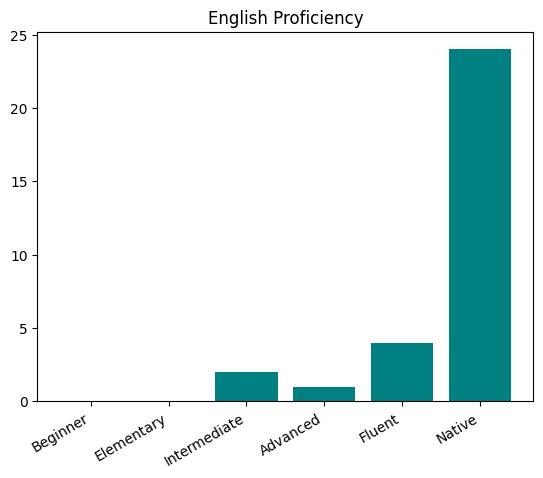}
         \label{fig: English Proficiency LLM}
     \end{subfigure}
     \hfill
     \begin{subfigure}{0.4\textwidth}
         \centering
         \includegraphics[width=\textwidth, height=6cm]{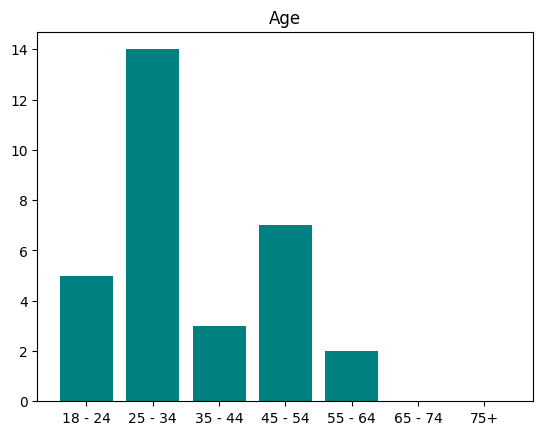}
         \label{Age LLM}
     \end{subfigure}
     \begin{subfigure}{0.4\textwidth}
         \centering
         \includegraphics[width=\textwidth, height=6cm]{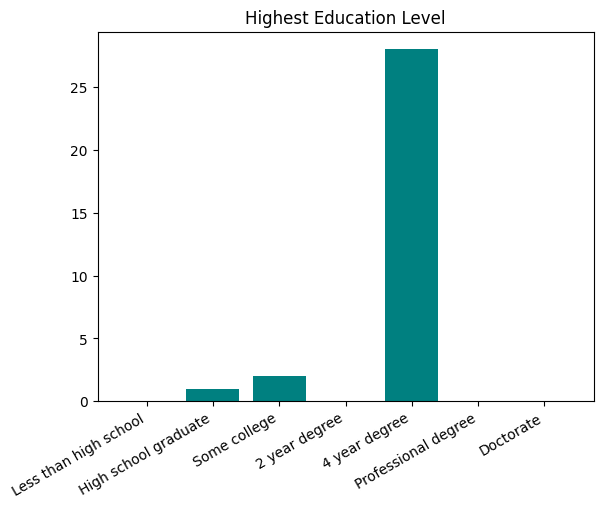}
         \label{Edu LLM}
     \end{subfigure}
     \hfill
     \begin{subfigure}{0.4\textwidth}
         \centering
         \includegraphics[width=\textwidth, height=6cm]{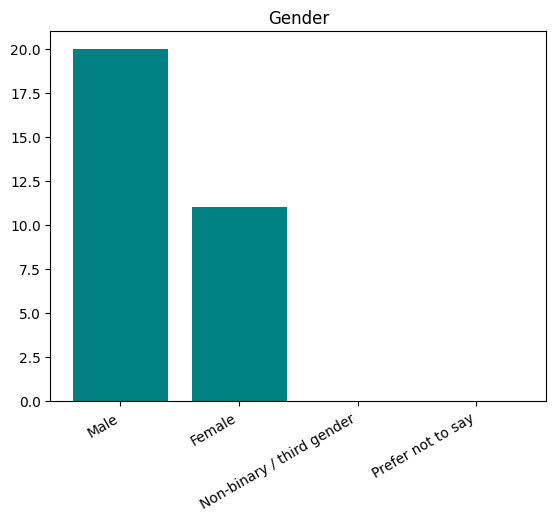}
         \label{Gender LLM}
     \end{subfigure}
        \caption{LLM Contexts Comparison Survey Demographics}
        \label{LLM Survey Demographics}
\end{figure}

\newpage

\subsection{Demographics of Participants of Summary Task 2}
\label{appendix:demographics_summary_survey2}
\begin{figure}[H]
     \centering
     \begin{subfigure}{.4\textwidth}
         \centering
         \includegraphics[width=\textwidth, height=6cm]{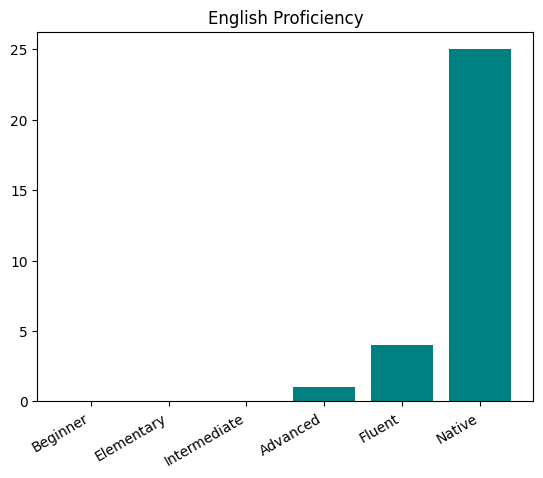}
         \label{fig: English Proficiency Sum 2}
     \end{subfigure}
     \hfill
     \begin{subfigure}{.4\textwidth}
         \centering
         \includegraphics[width=\textwidth, height=6cm]{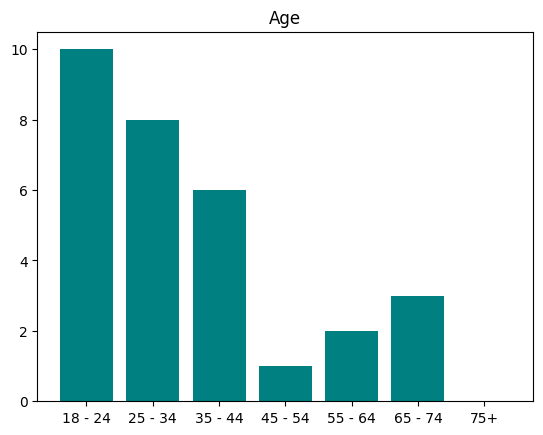}
         \label{Age Sum 2}
     \end{subfigure}
     \vskip\baselineskip
     \begin{subfigure}{.4\textwidth}
         \centering
         \includegraphics[width=\textwidth, height=6cm]{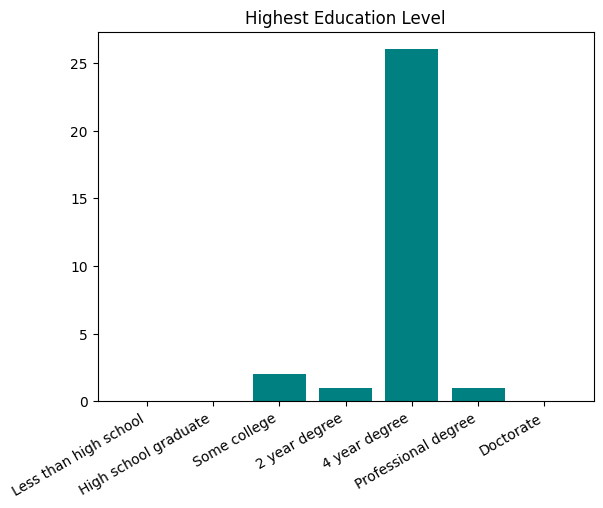}
         \label{Edu Sum 2}
     \end{subfigure}
     \hfill
     \begin{subfigure}{.4\textwidth}
         \centering
         \includegraphics[width=\textwidth, height=6cm]{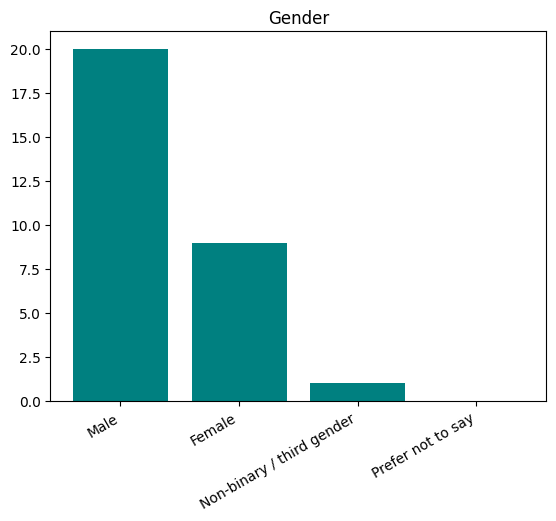}
         \label{Gender Sum 2}
     \end{subfigure}
        \caption{Summary Survey 2 Demographics}
        \label{Sum 2 Demographics}
\end{figure}

\newpage

\subsection{Demographics of HSE dataset Annotators}
\label{appendix:demographics_ground_truth}
\begin{figure}[H]
     \centering
     \includegraphics[width=1\textwidth]{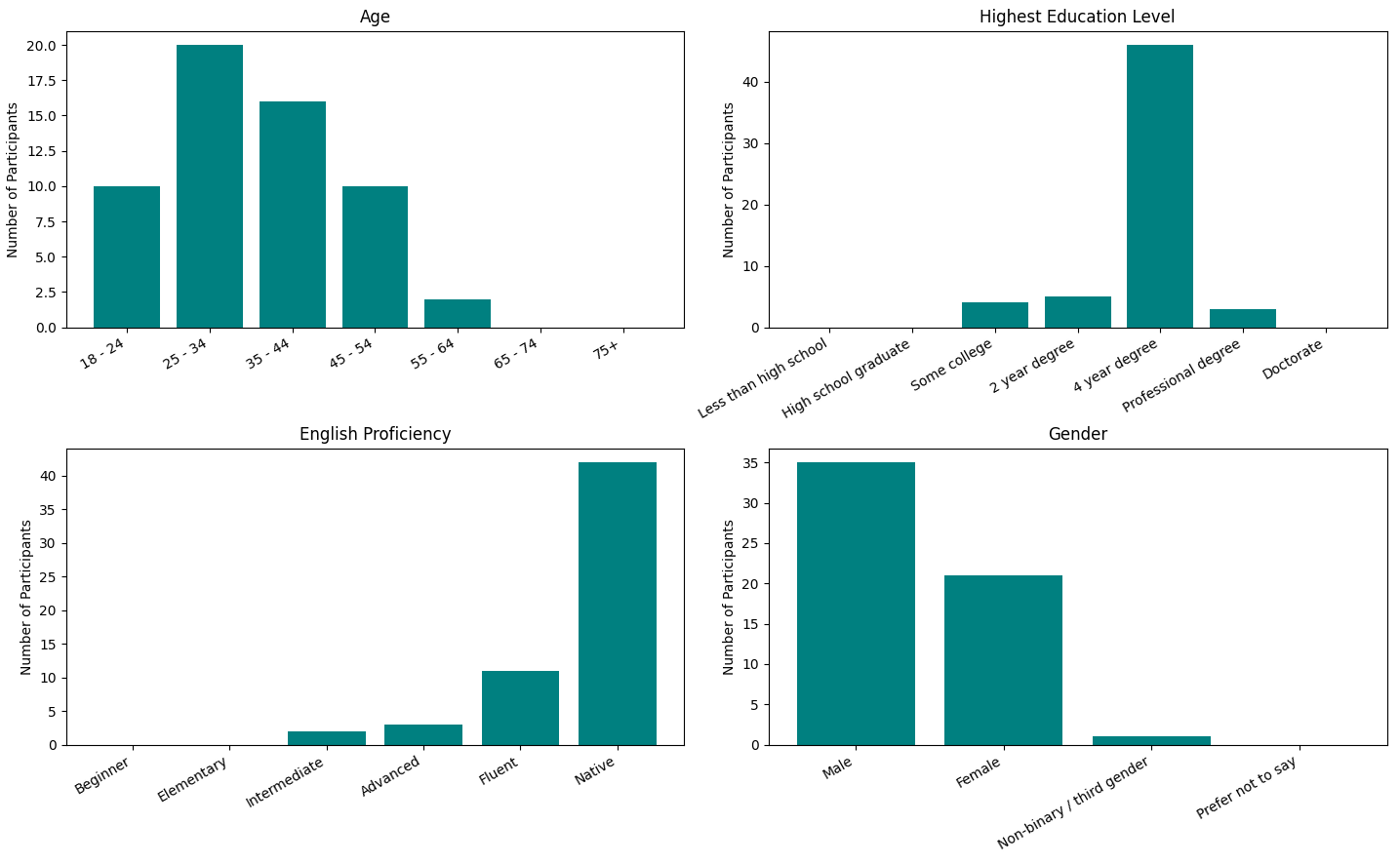}
    \caption{Crowd Source Survey Demographics}
    \label{Crowd Source Survey Demographics}
\end{figure}

\section{Screenshots of surveys}
\label{appendix:survey_scnshots}

\subsection{Comparison of Explicit and Implicit Contextualisation Survey}
\begin{figure}[H]
    \centering
    \includegraphics[width=.8\textwidth]{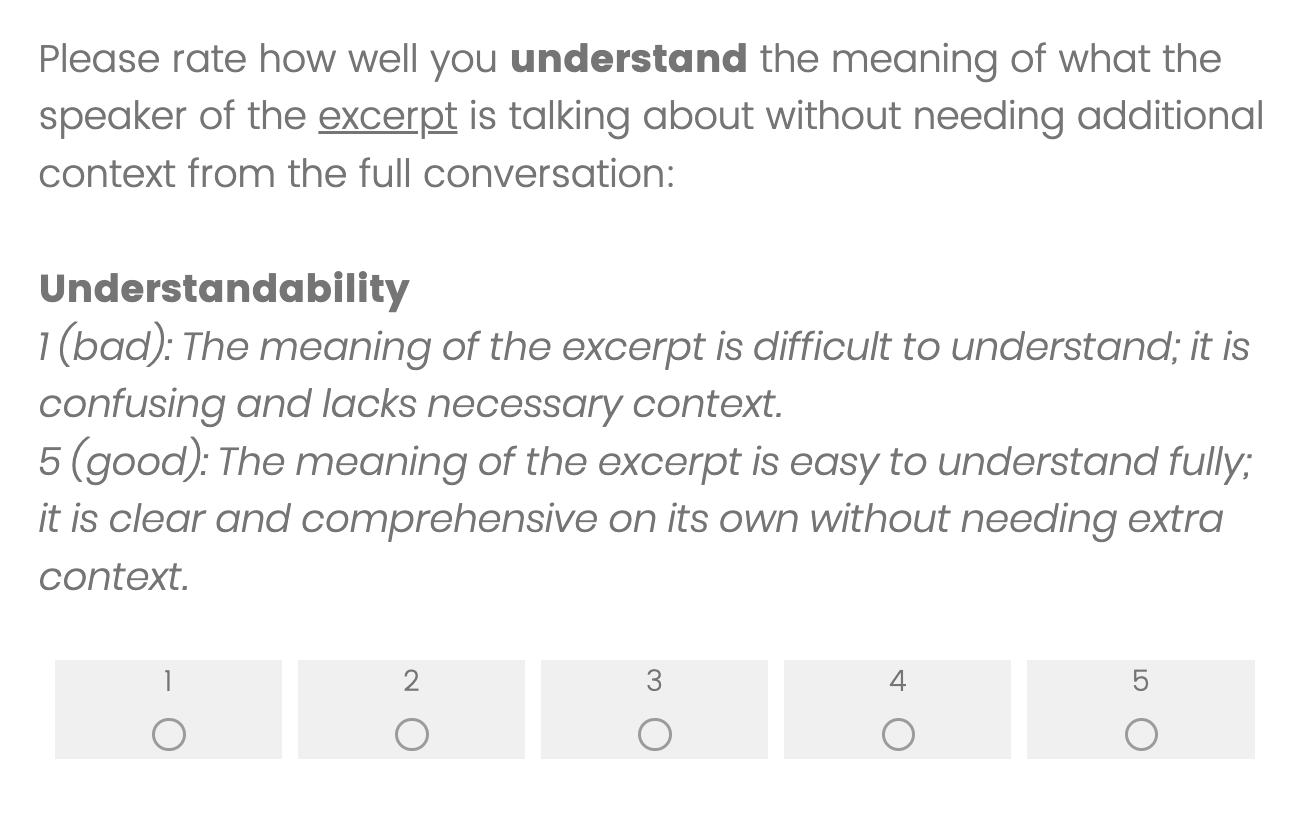}
    \caption{Understandability Question}
    \label{Figure ##: Understandability Question}
\end{figure}
\begin{figure}[H]
    \centering
    \includegraphics[width=.8\textwidth]{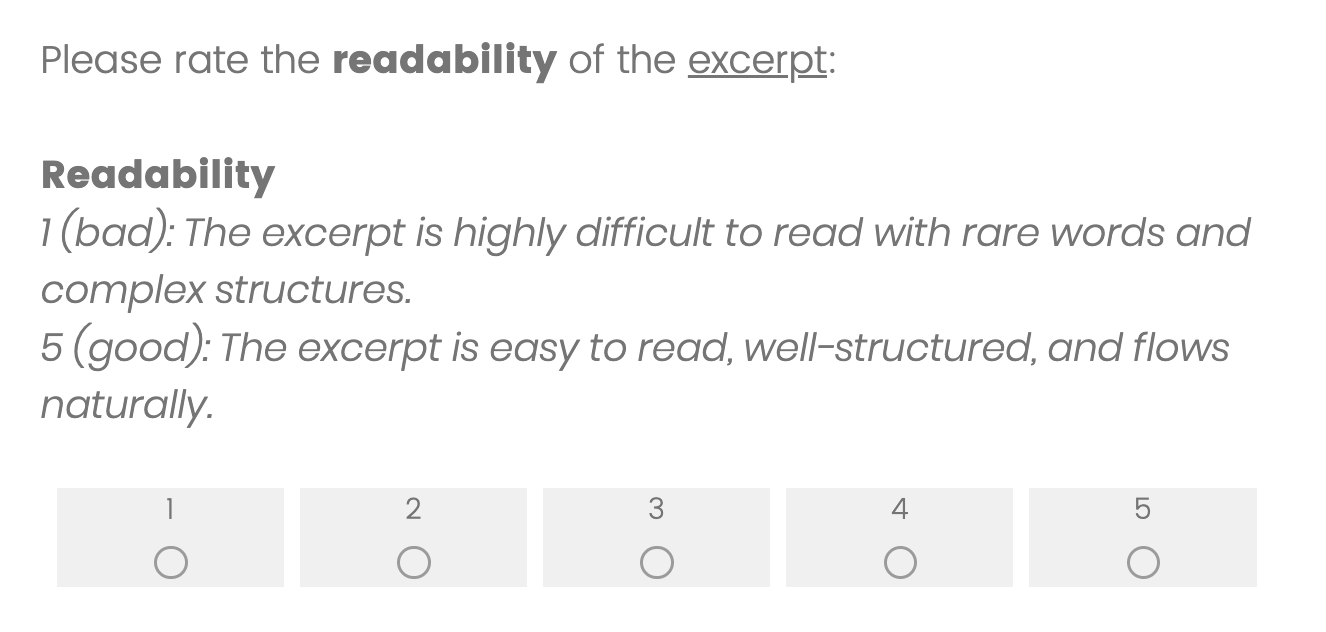}
    \caption{Readability Question}
    \label{Figure ##: Readability Question}
\end{figure}
\begin{figure}[H]
    \centering
    \includegraphics[width=.8\textwidth]{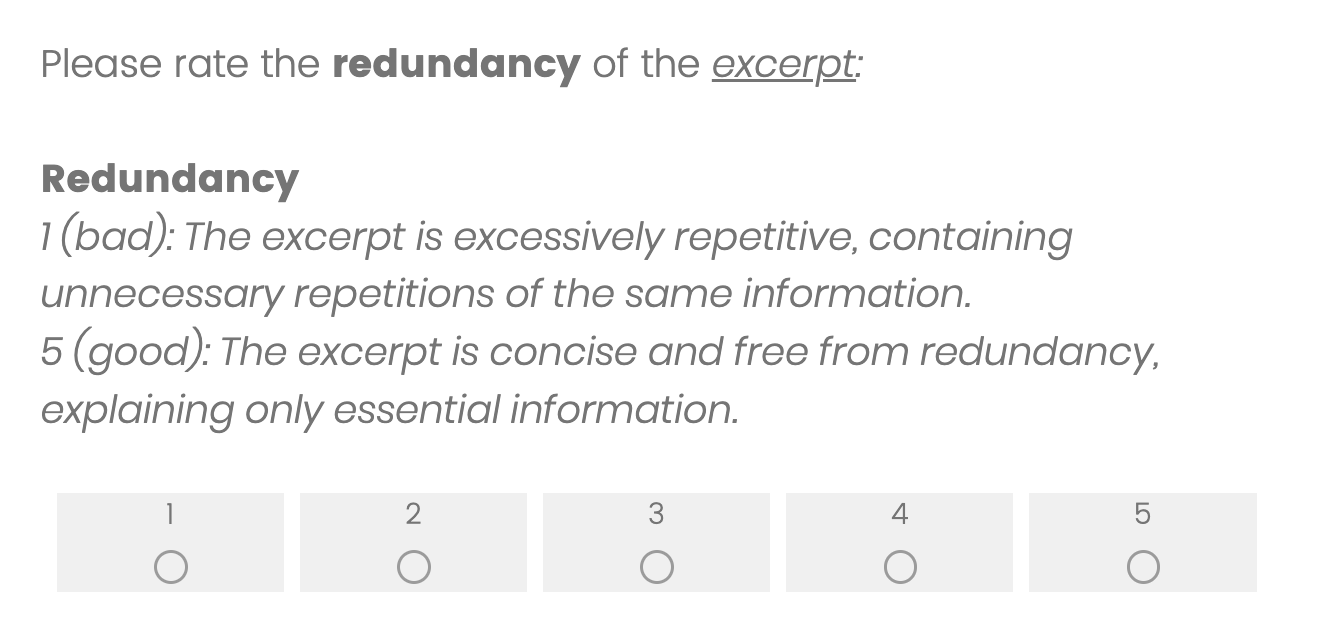}
    \caption{Redundancy Question}
    \label{Figure ##: Redundancy Question}
\end{figure}
\begin{figure}[H]
    \centering
    \includegraphics[width=.8\textwidth]{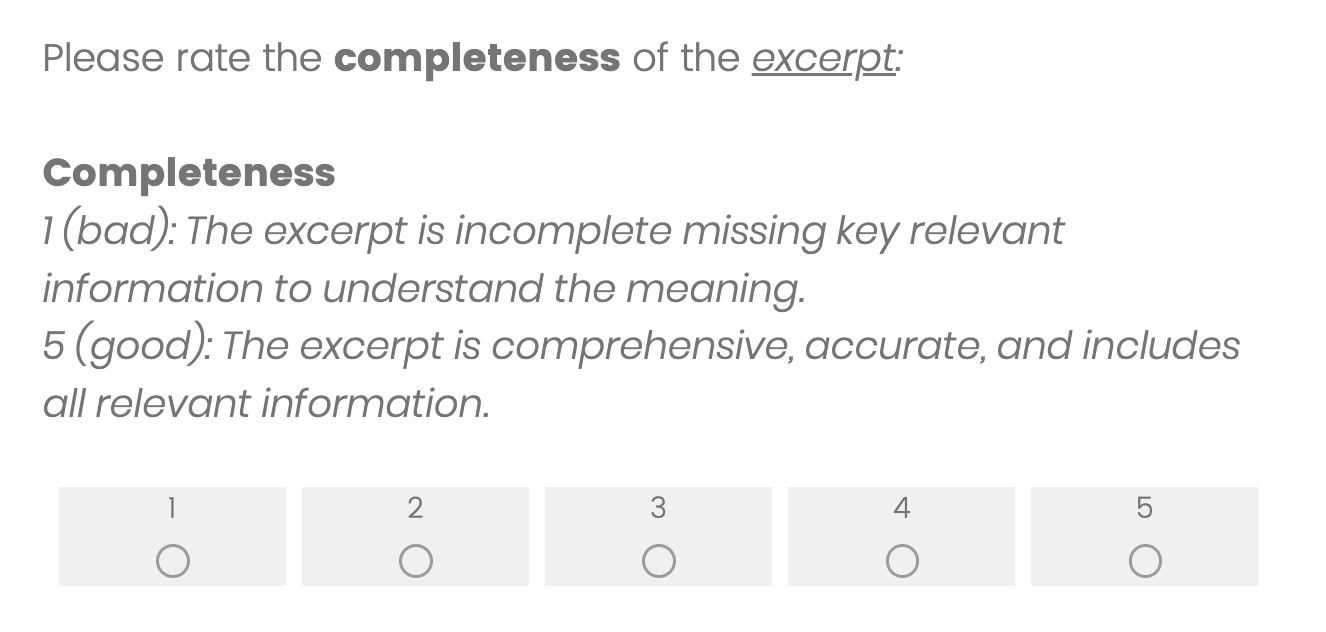}
    \caption{Completeness Question}
    \label{Figure ##: Completeness Question}
\end{figure}
\begin{figure}[H]
    \centering
    \includegraphics[width=.8\textwidth]{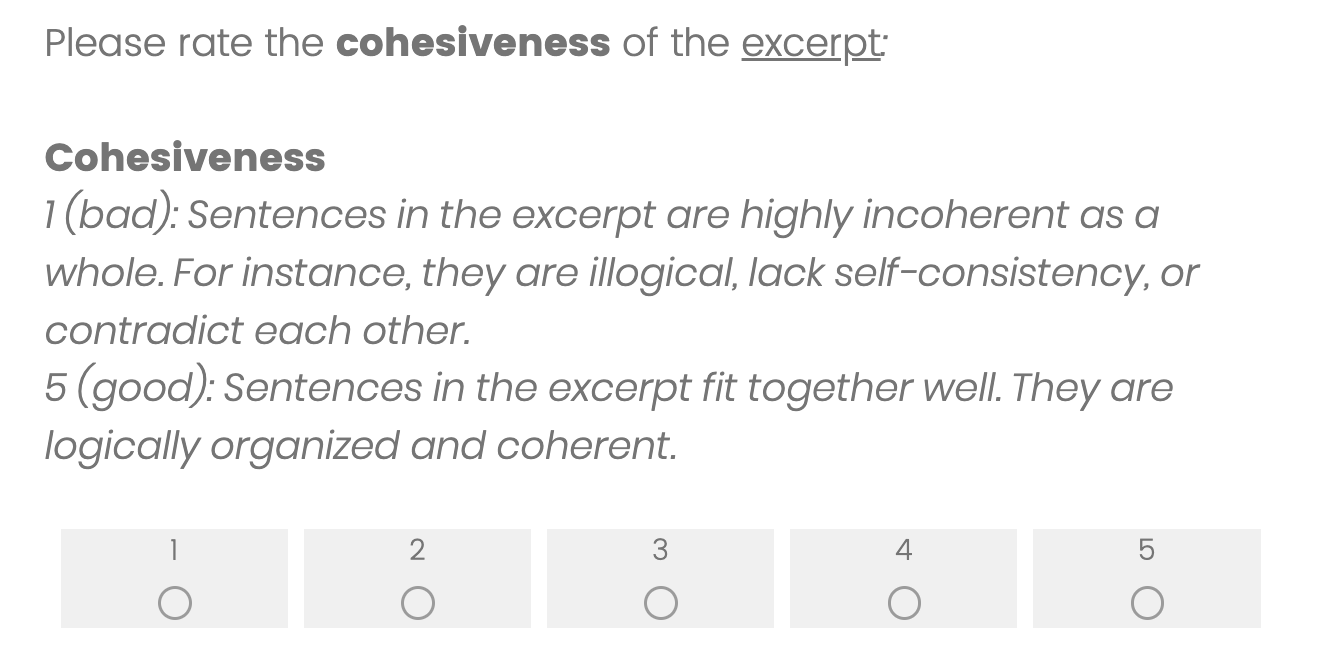}
    \caption{Cohesiveness Question}
    \label{Figure ##: Cohesiveness Question}
\end{figure}
\begin{figure}[H]
    \centering
    \includegraphics[width=.8\textwidth]{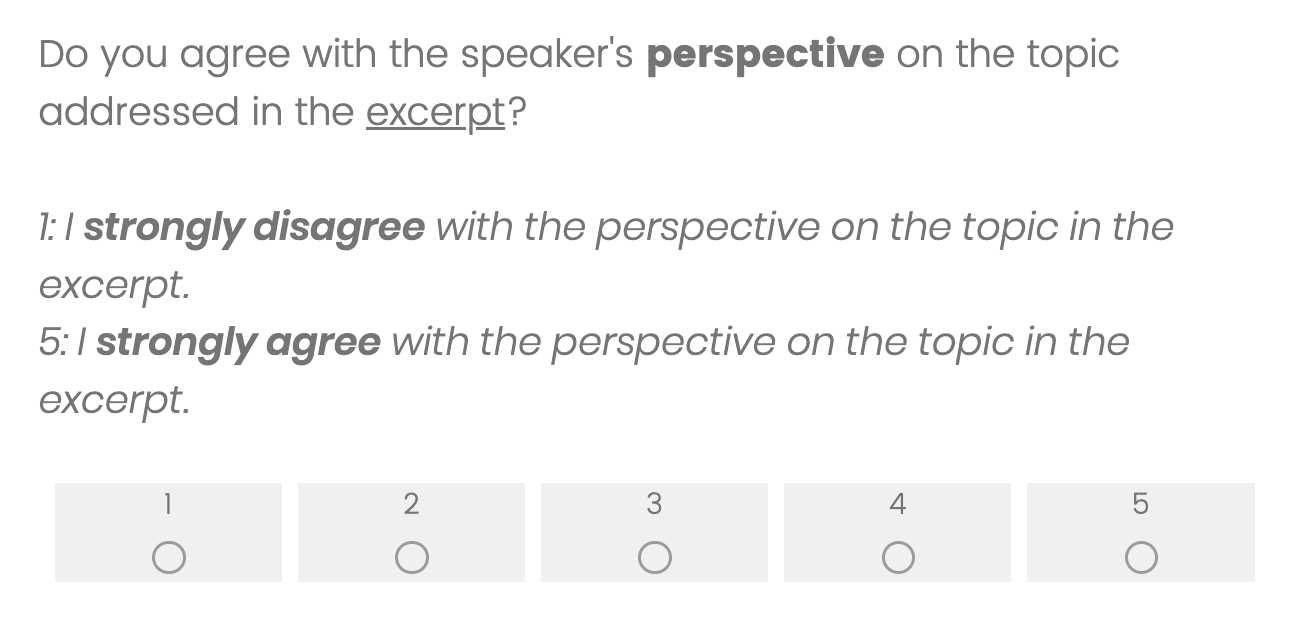}
    \caption{Perspective Question}
    \label{Figure ##: Perspective Question}
\end{figure}
\begin{figure}[H]
    \centering
    \includegraphics[width=.8\textwidth]{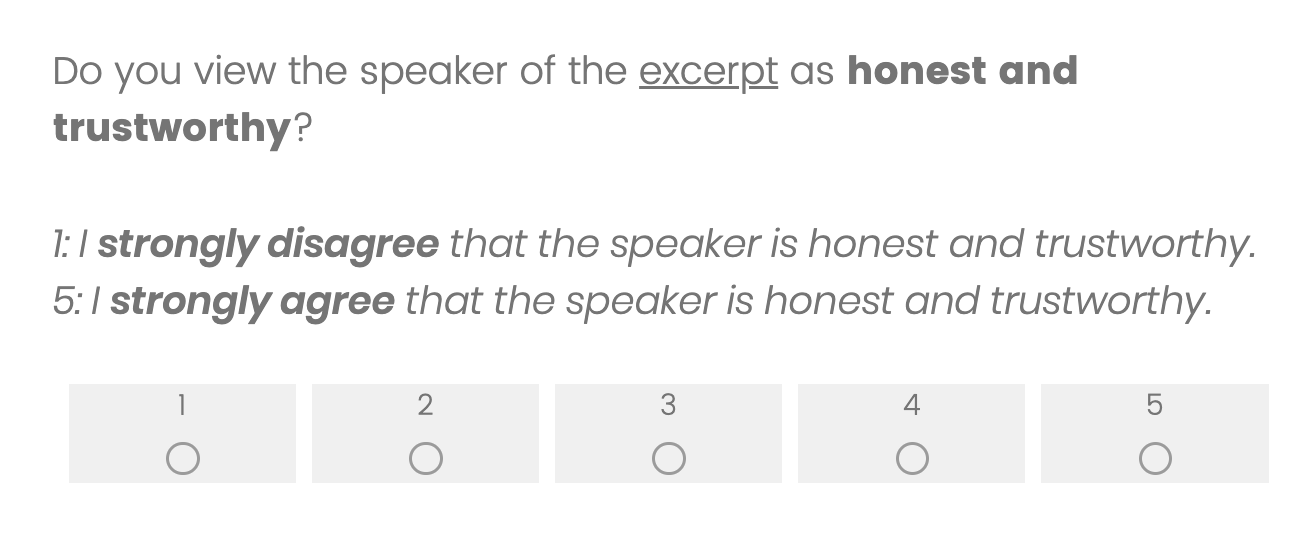}
    \caption{Honest and Trustworthy Question}
    \label{Figure ##: Honest and Trustworthy Question}
\end{figure}
\begin{figure}[H]
    \centering
    \includegraphics[width=.8\textwidth]{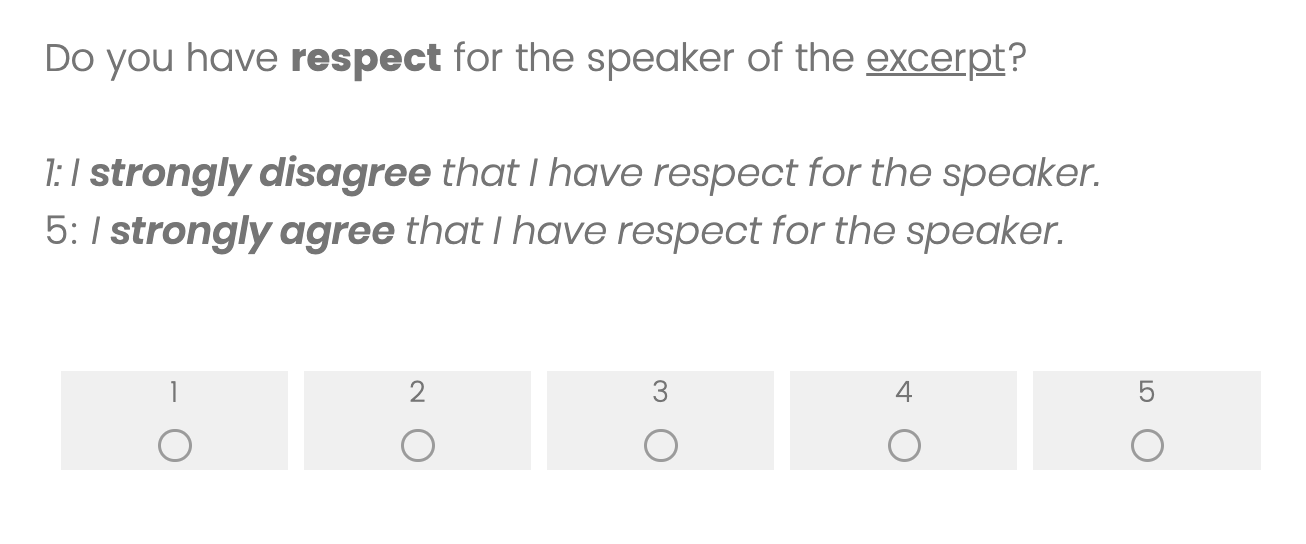}
    \caption{Respect Question}
    \label{Figure ##: Respect Question}
\end{figure}
\begin{figure}[H]
    \centering
    \includegraphics[width=.8\textwidth]{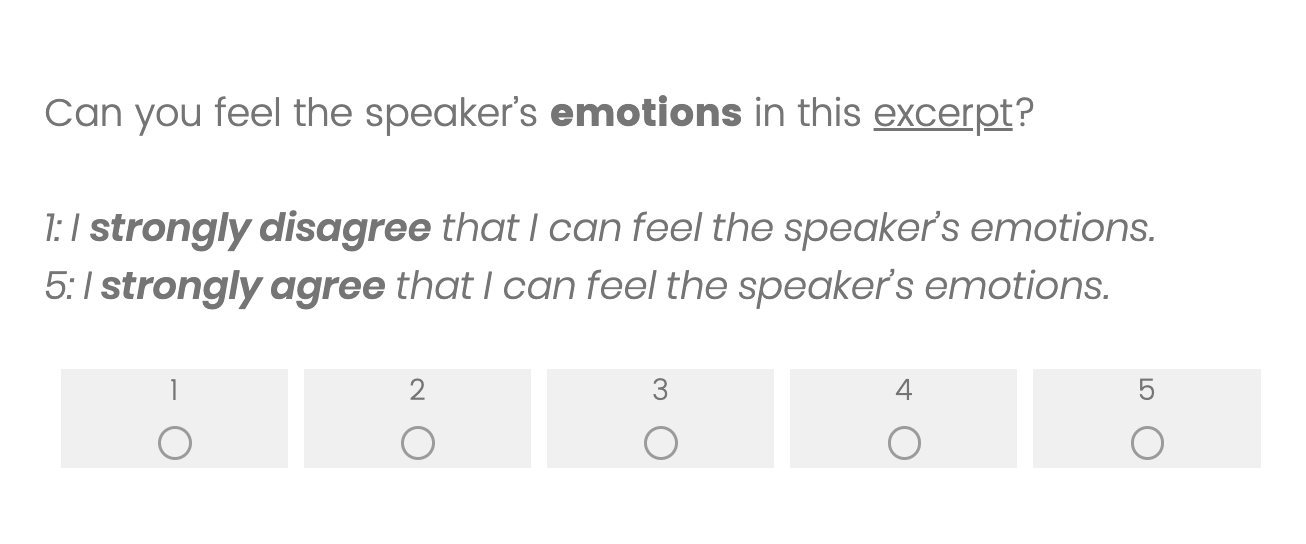}
    \caption{Empathy Question}
    \label{Figure ##: Empathy Question}
\end{figure}
\begin{figure}[H]
    \centering
    \includegraphics[width=.8\textwidth]{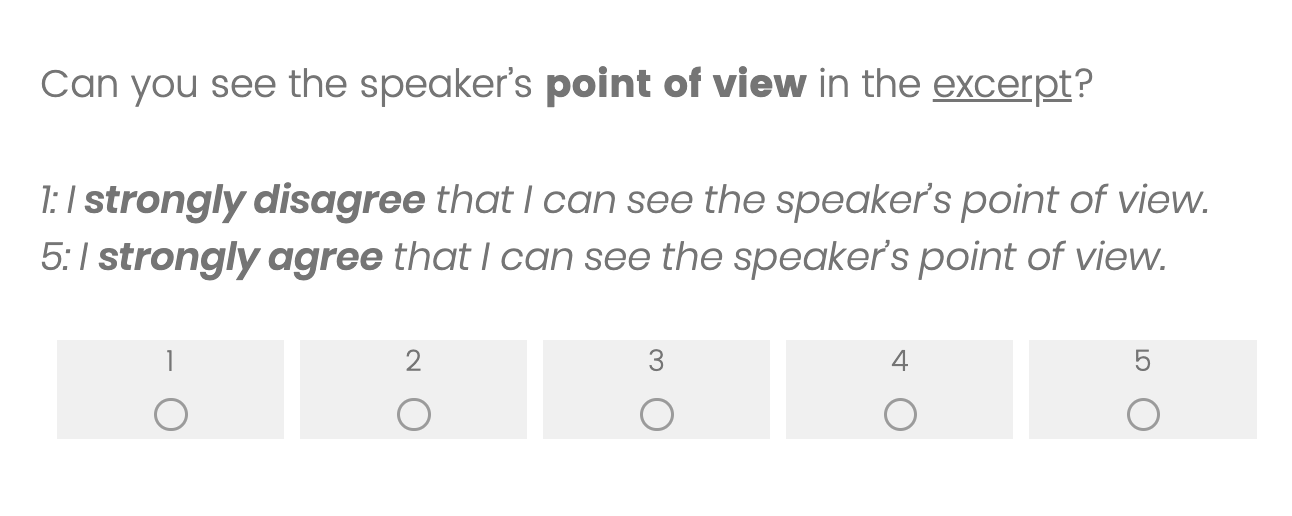}
    \caption{POV Question}
    \label{Figure ##: POV Question}
\end{figure}
\begin{figure}[H]
    \centering
    \includegraphics[width=.8\textwidth]{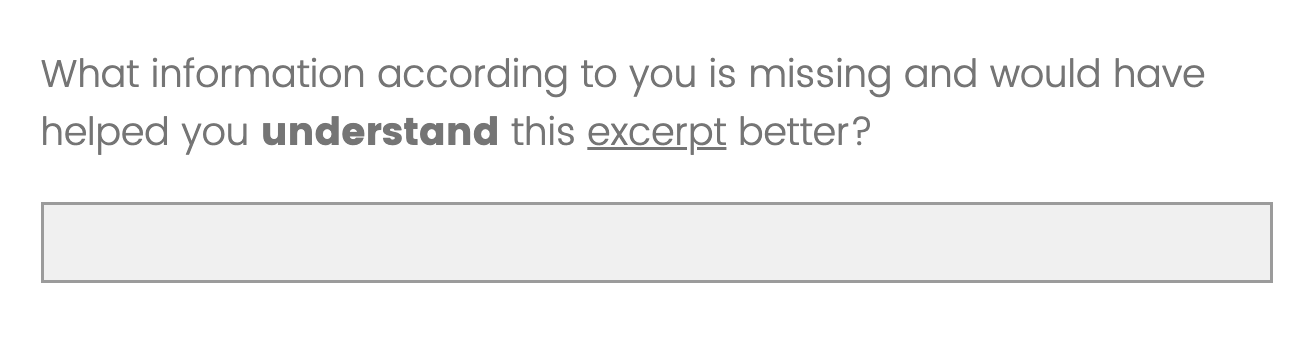}
    \caption{Missing Info Question}
    \label{Figure ##: Missing Info Question}
\end{figure}
\begin{figure}[H]
    \centering
    \includegraphics[width=.8\textwidth]{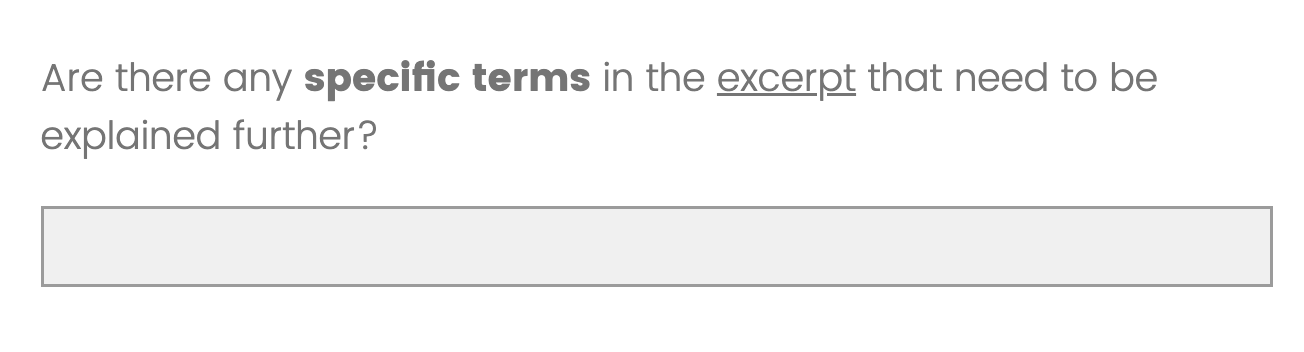}
    \caption{Clarify Terms Question}
    \label{Figure ##: Clarify Terms Question}
\end{figure}
\begin{figure}[H]
    \centering
    \includegraphics[width=.8\textwidth]{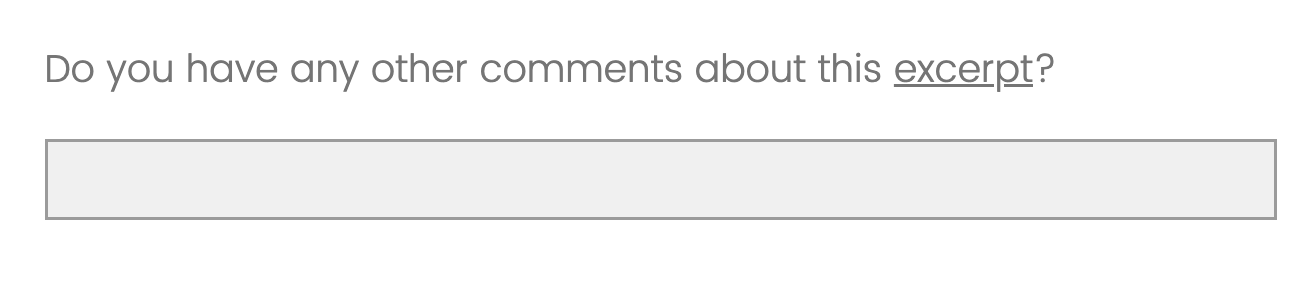}
    \caption{Comments Question}
    \label{Figure ##: Comments Question}
\end{figure}
\begin{figure}[H]
    \centering
    \includegraphics[width=.8\textwidth]{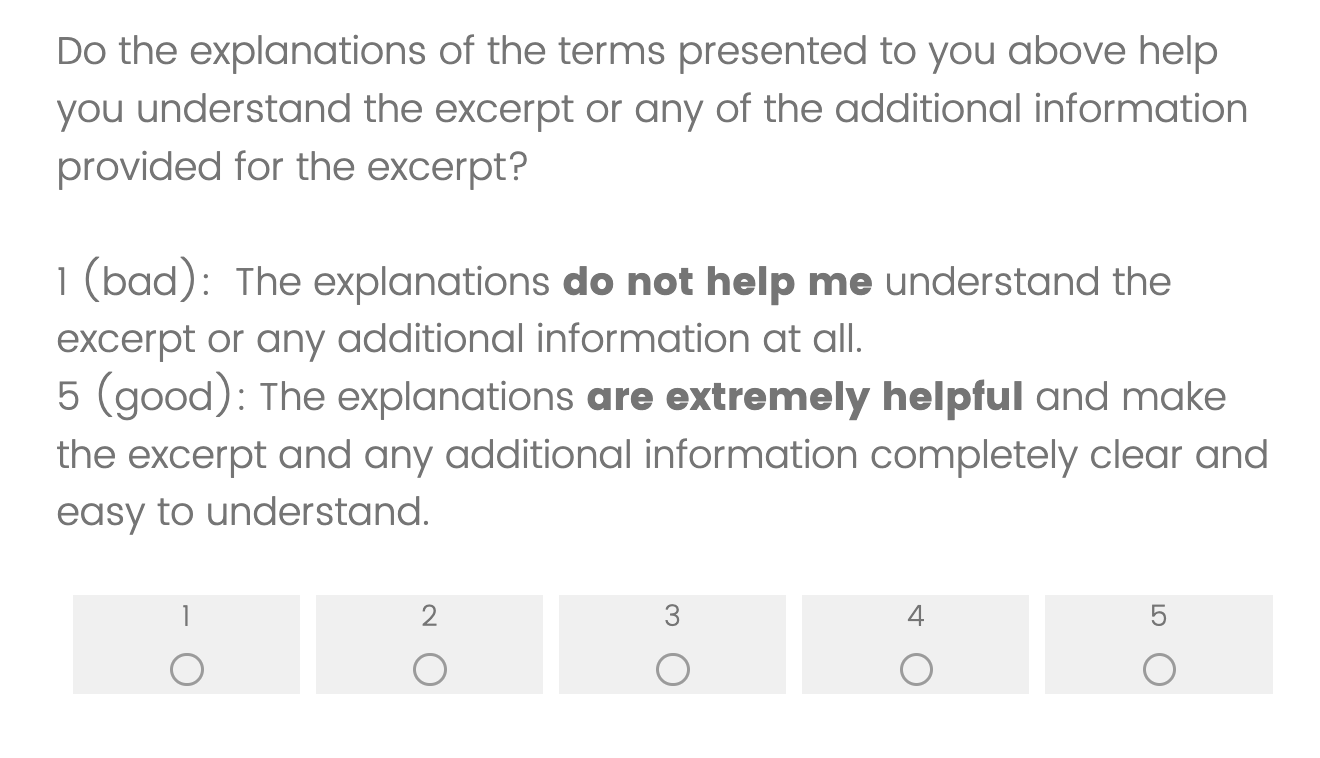}
    \caption{ICL Explanation of Terms Question}
    \label{Figure ##: ICL Explanation of Termss Question}
\end{figure}
\begin{figure}[H]
    \centering
    \includegraphics[width=.8\textwidth]{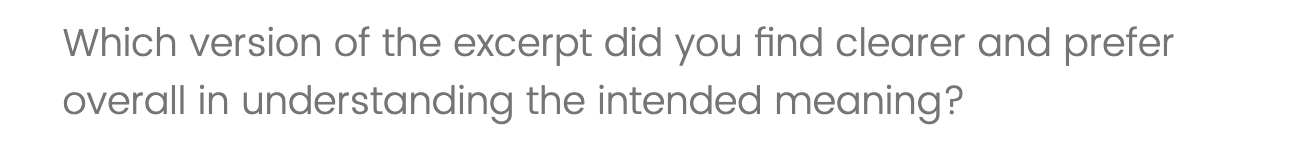}
    \caption{Rank Contexts Question}
    \label{Figure ##: Rank Contexts Question}
\end{figure}
\begin{figure}[H]
    \centering
    \includegraphics[width=.8\textwidth]{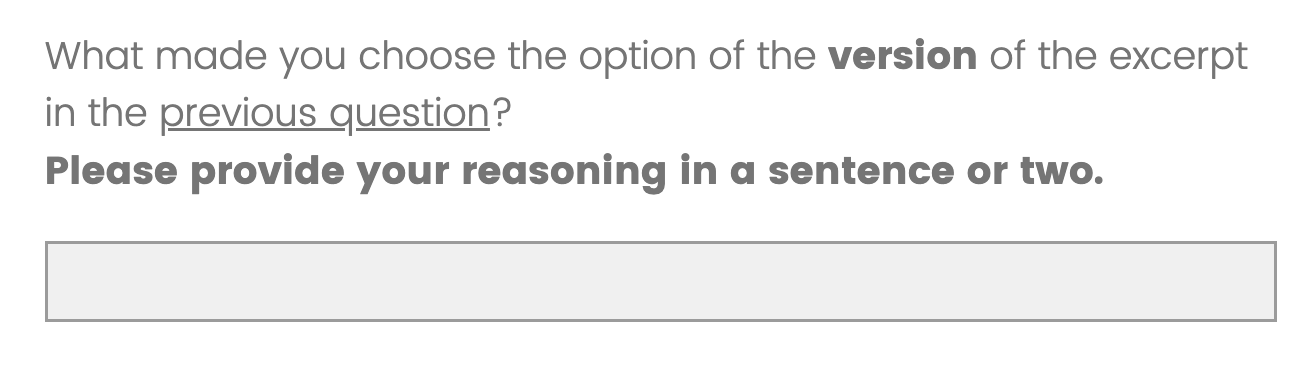}
    \caption{Justify Rank Question}
    \label{Figure ##: Justify Rank Question}
\end{figure}

\subsection{Crowd-Sourced Annotations for HSE}
\label{appendix:hse_annotations}
\begin{figure}[H]
    \centering
    \includegraphics[width=.8\textwidth]{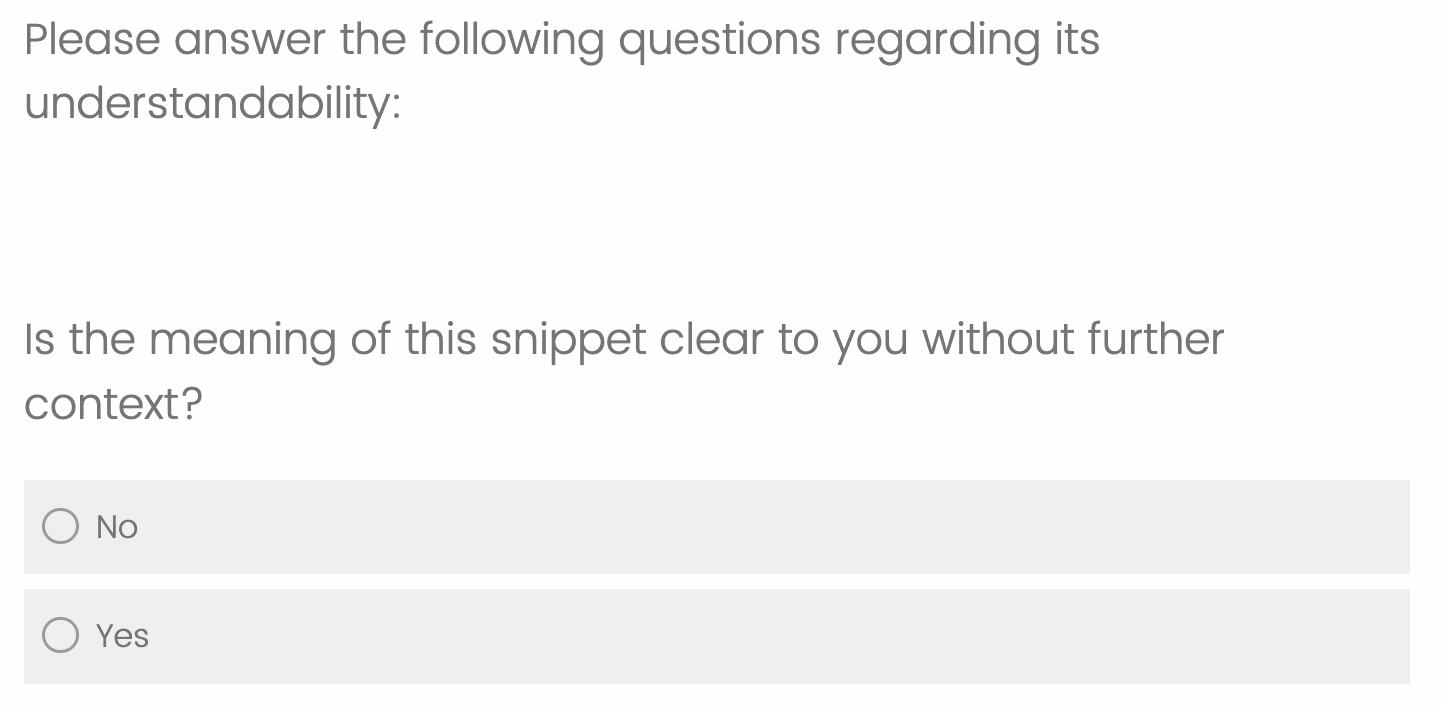}
    \caption{Understandability Question}
    \label{Figure ##: Understandability YN Question}
\end{figure}
\begin{figure}[H]
    \centering
    \includegraphics[width=.8\textwidth]{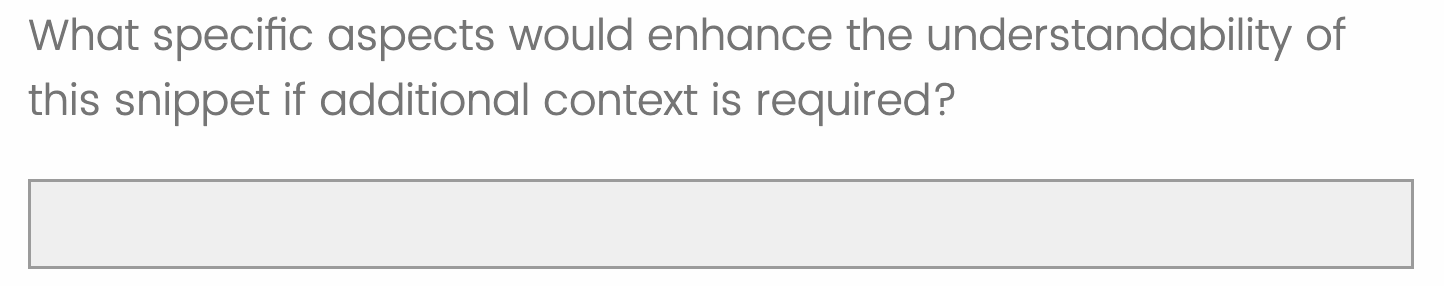}
    \caption{Enhance Understandability Question}
    \label{Figure ##: Enhance Understandability Question}
\end{figure}
\begin{figure}[H]
    \centering
    \includegraphics[width=.8\textwidth]{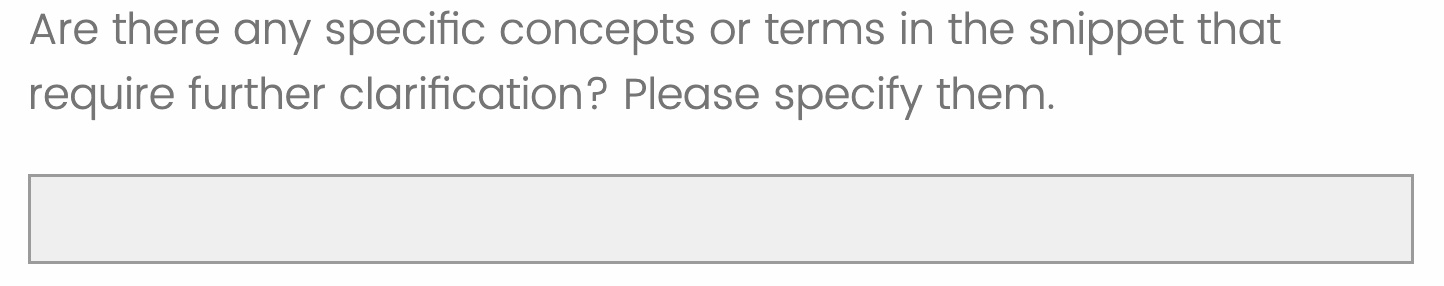}
    \caption{Further Clarification Question}
    \label{Figure ##: Further Clarification Question}
\end{figure}
\begin{figure}[H]
    \centering
    \includegraphics[width=.8\textwidth]{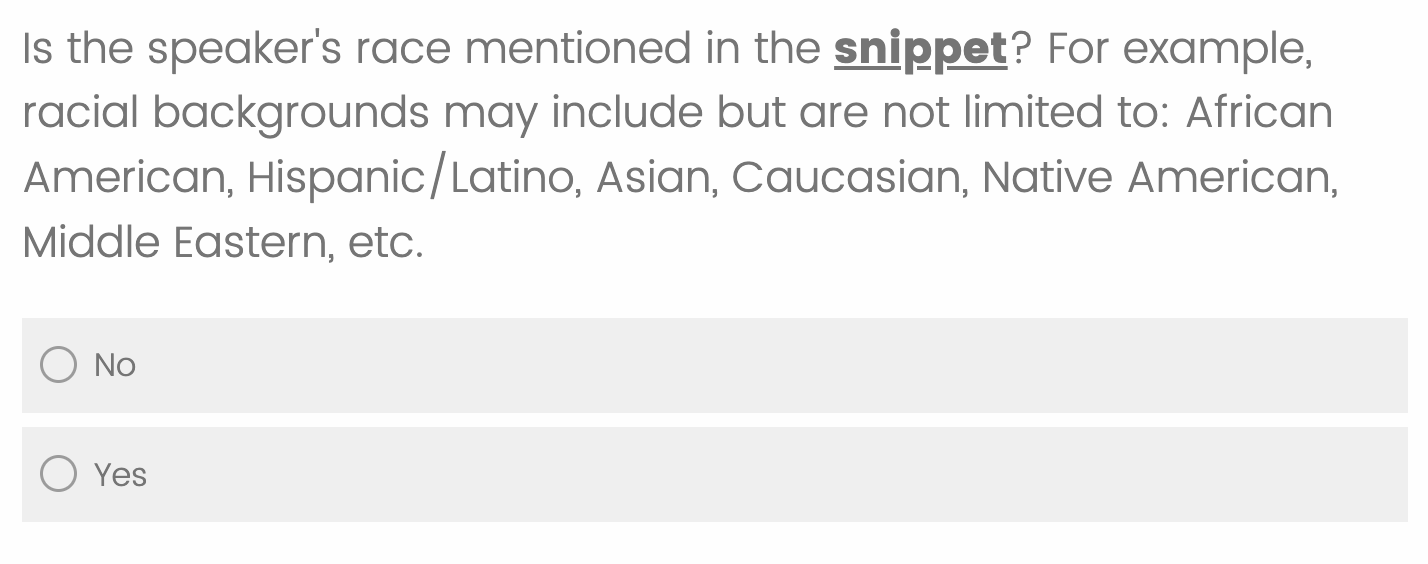}
    \caption{Speaker Race Question}
    \label{Figure ##: Speaker Race YN Question}
\end{figure}
\begin{figure}[H]
    \centering
    \includegraphics[width=.8\textwidth]{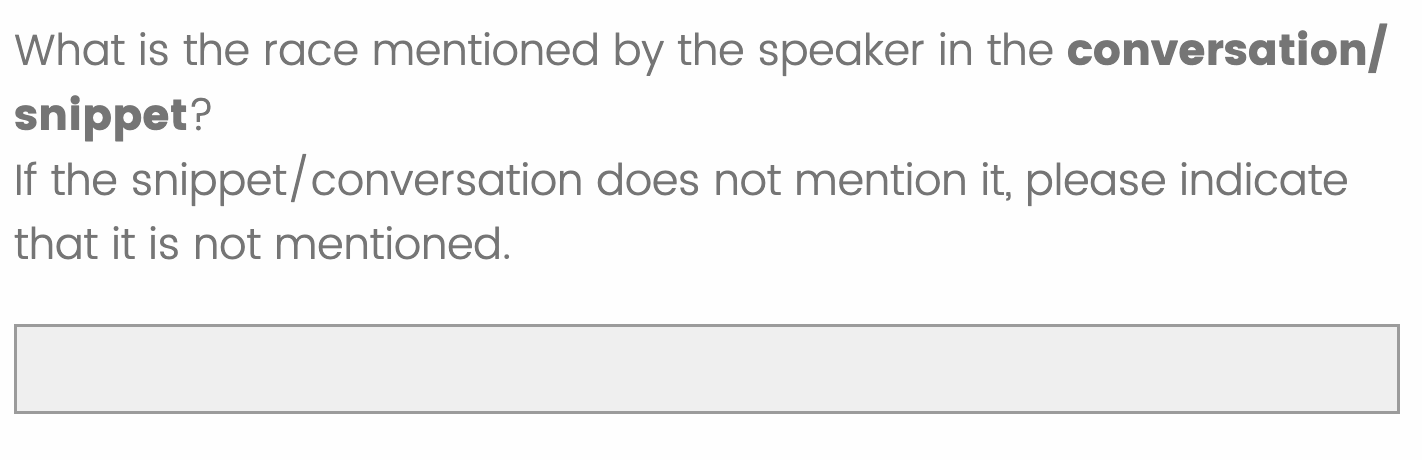}
    \caption{Speaker Race Text Entry Question}
    \label{Figure ##: Speaker Race Text Entry Question}
\end{figure}
\begin{figure}[H]
    \centering
    \includegraphics[width=.8\textwidth]{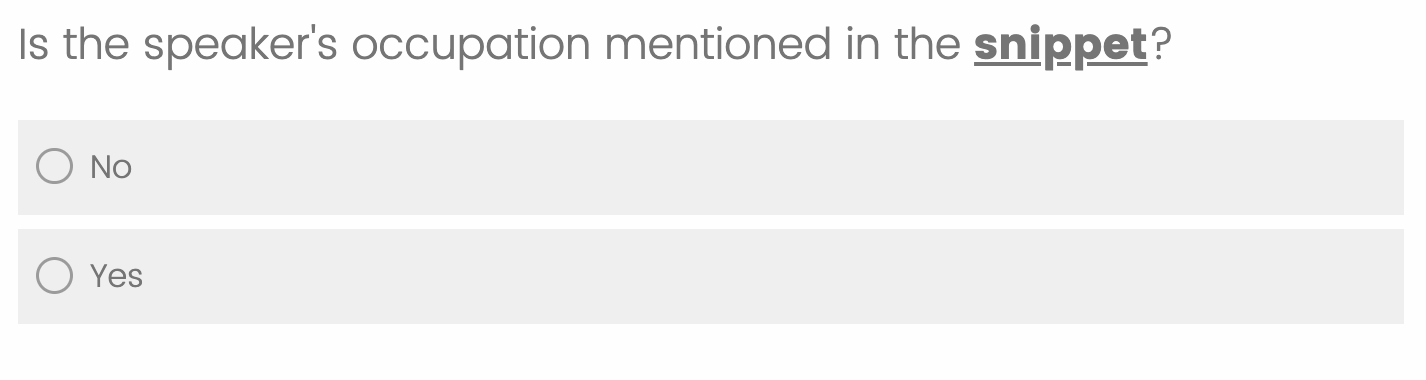}
    \caption{Speaker Occupation Question}
    \label{Figure ##: Speaker Occupation YN Question}
\end{figure}
\begin{figure}[H]
    \centering
    \includegraphics[width=.8\textwidth]{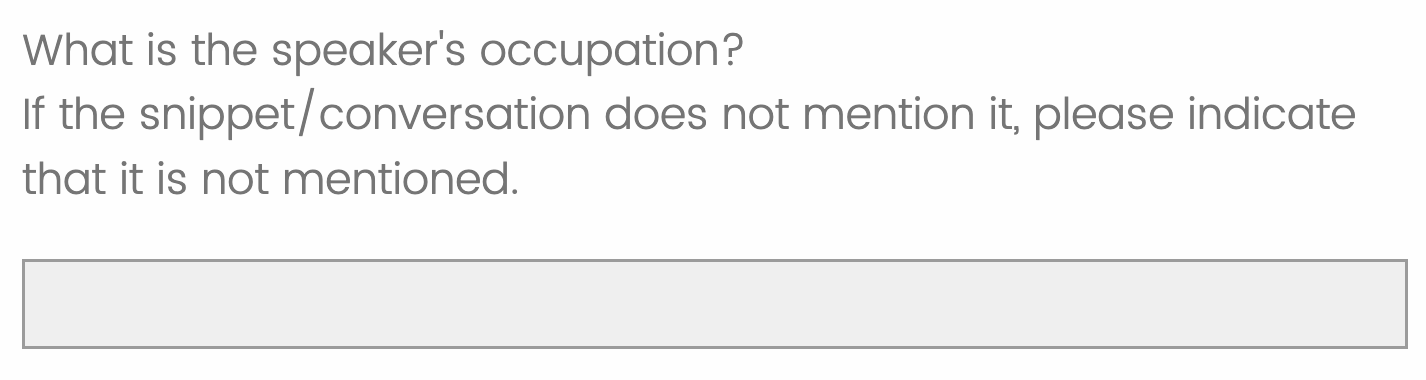}
    \caption{Speaker Occupation Text Entry Question}
    \label{Figure ##: Speaker Occupation Text Entry Question}
\end{figure}
\begin{figure}[H]
    \centering
    \includegraphics[width=.8\textwidth]{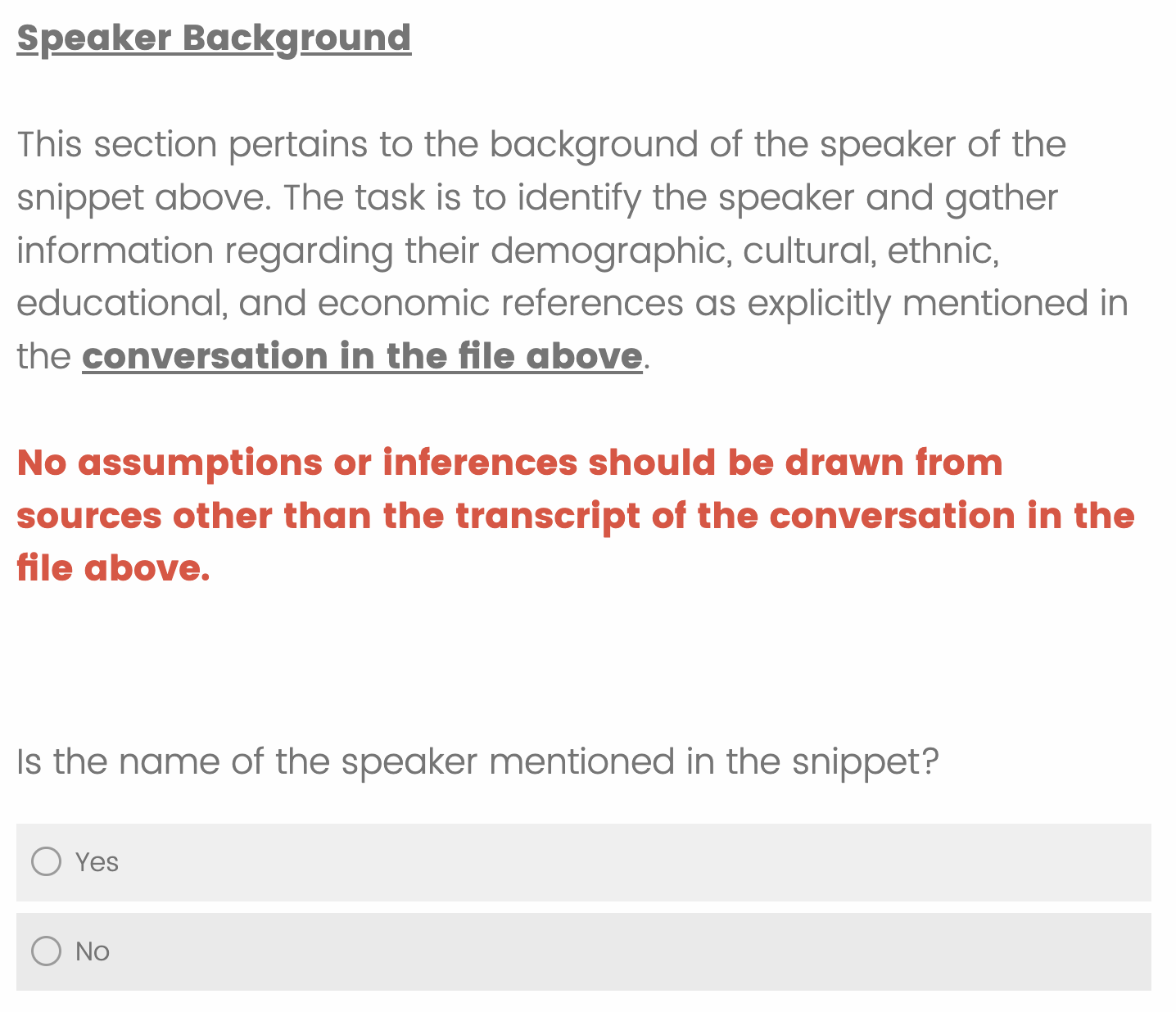}
    \caption{Speaker Name Question}
    \label{Figure ##: Speaker Name YN Question}
\end{figure}
\begin{figure}[H]
    \centering
    \includegraphics[width=.8\textwidth]{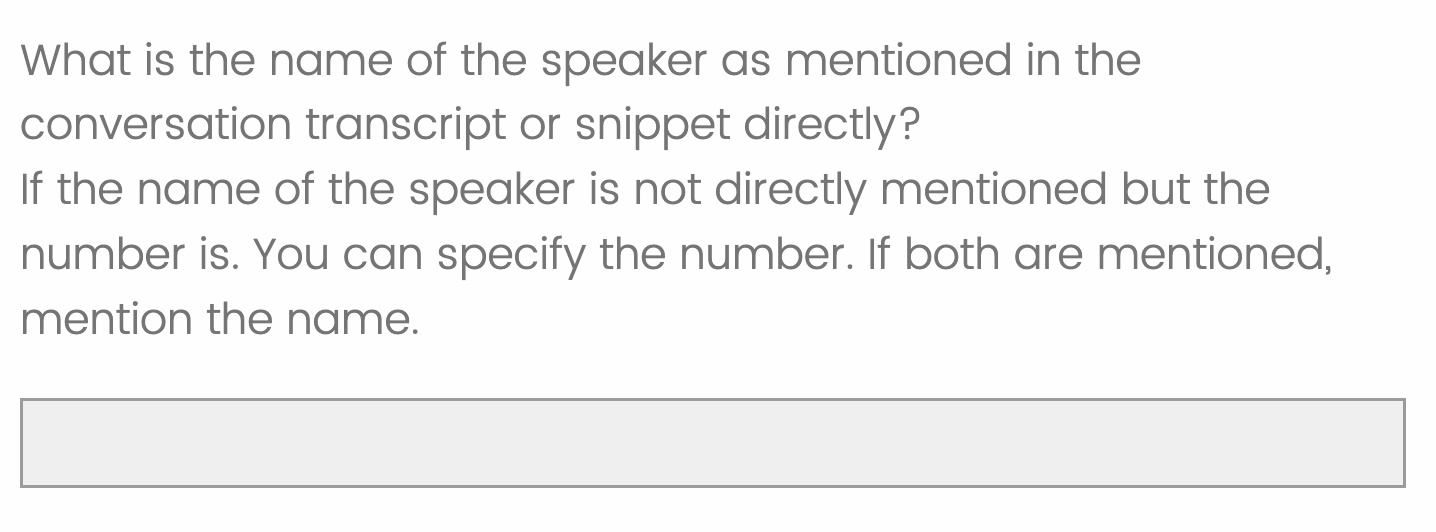}
    \caption{Speaker Name Text Entry Question}
    \label{Figure ##: Speaker Name Text Entry Question}
\end{figure}
\begin{figure}[H]
    \centering
    \includegraphics[width=.8\textwidth]{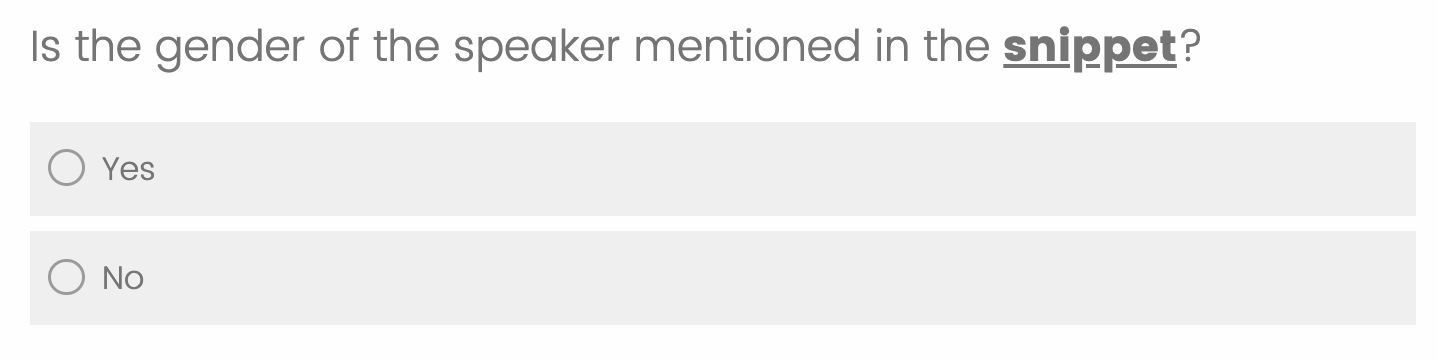}
    \caption{Gender Question}
    \label{Figure ##: Gender YN Question}
\end{figure}
\begin{figure}[H]
    \centering
    \includegraphics[width=.8\textwidth]{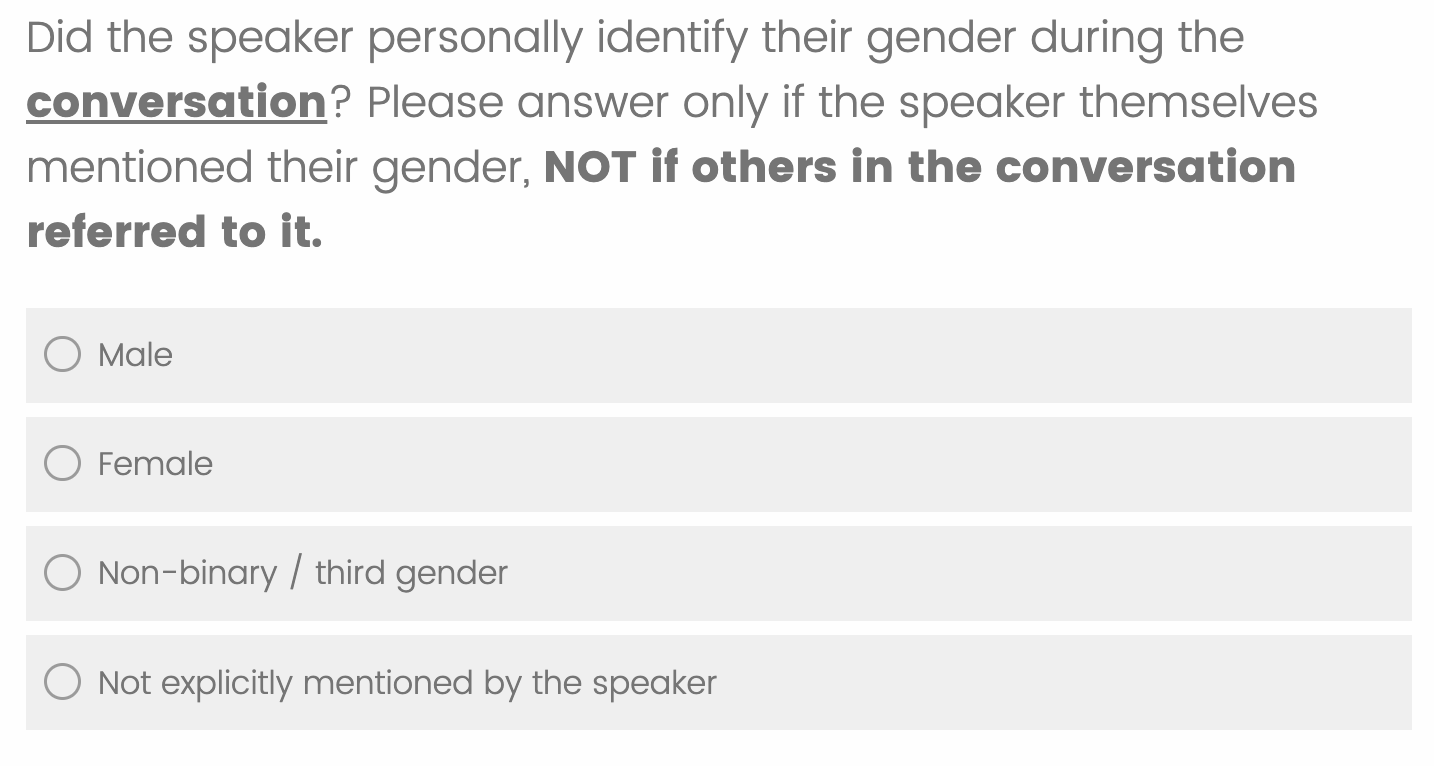}
    \caption{Speaker Gender Select Question}
    \label{Figure ##: Speaker Gender Select Question}
\end{figure}
\begin{figure}[H]
    \centering
    \includegraphics[width=.8\textwidth]{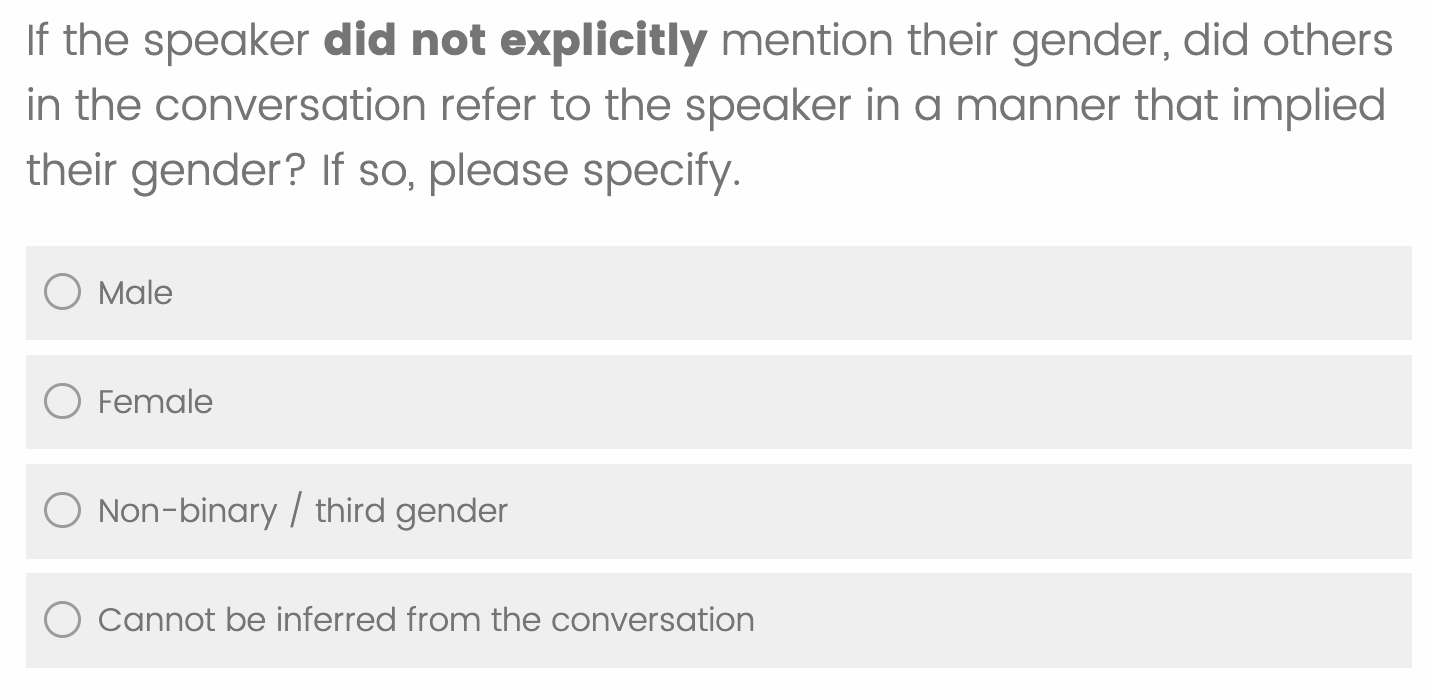}
    \caption{Speaker Gender Select 2 Question}
    \label{Figure ##: Speaker Gender Select 2 Question}
\end{figure}
\begin{figure}[H]
    \centering
    \includegraphics[width=.8\textwidth]{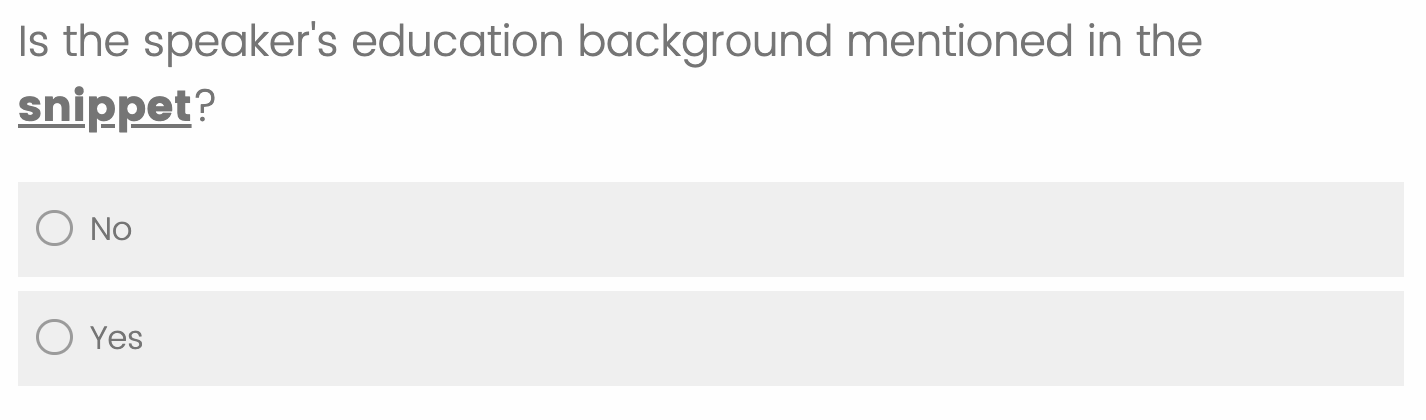}
    \caption{Speaker Education Question}
    \label{Figure ##: Speaker Education YN Question}
\end{figure}
\begin{figure}[H]
    \centering
    \includegraphics[width=.8\textwidth]{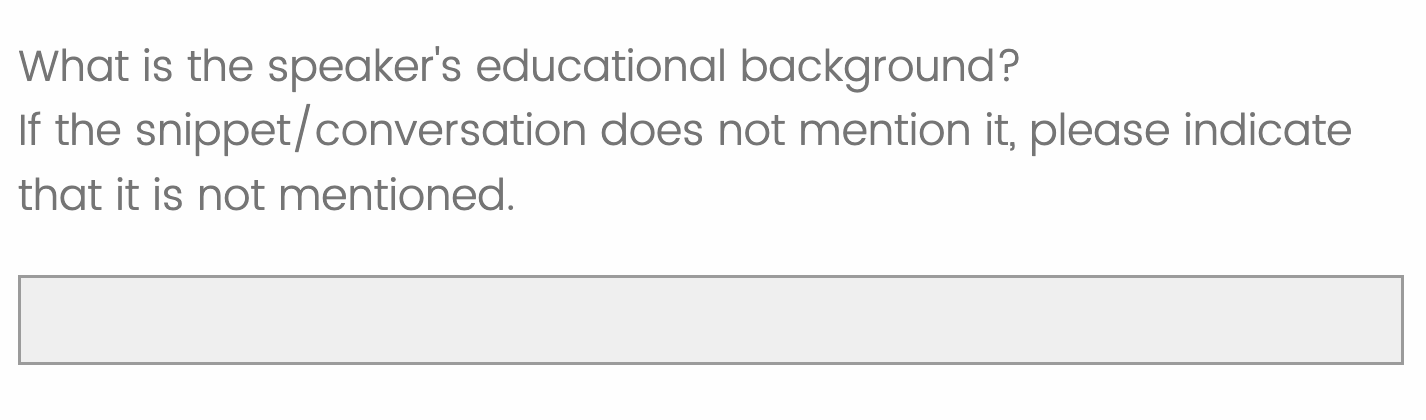}
    \caption{Speaker Education Text Entry Question}
    \label{Figure ##: Speaker Education Text Entry Question}
\end{figure}
\begin{figure}[H]
    \centering
    \includegraphics[width=.8\textwidth]{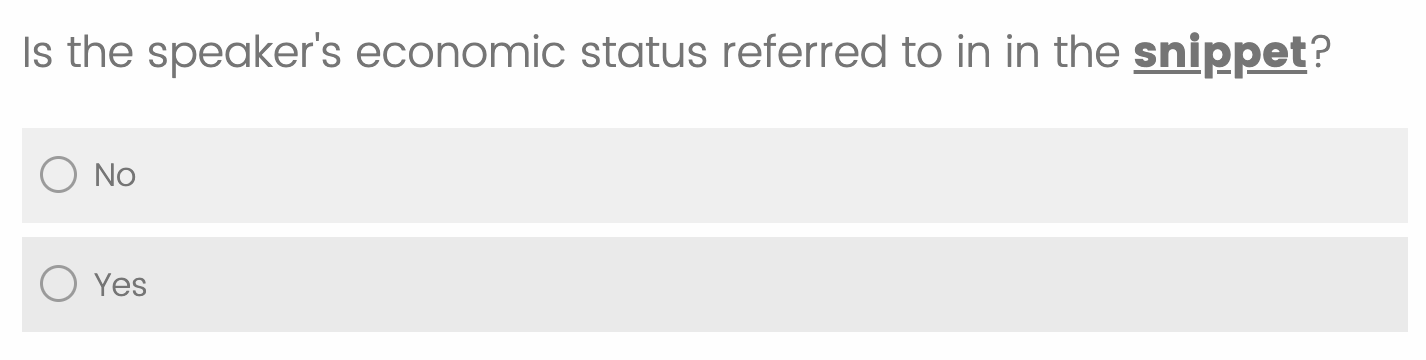}
    \caption{Speaker Economic Status Question}
    \label{Figure ##: Speaker Economic Status YN Question}
\end{figure}
\begin{figure}[H]
    \centering
    \includegraphics[width=.8\textwidth]{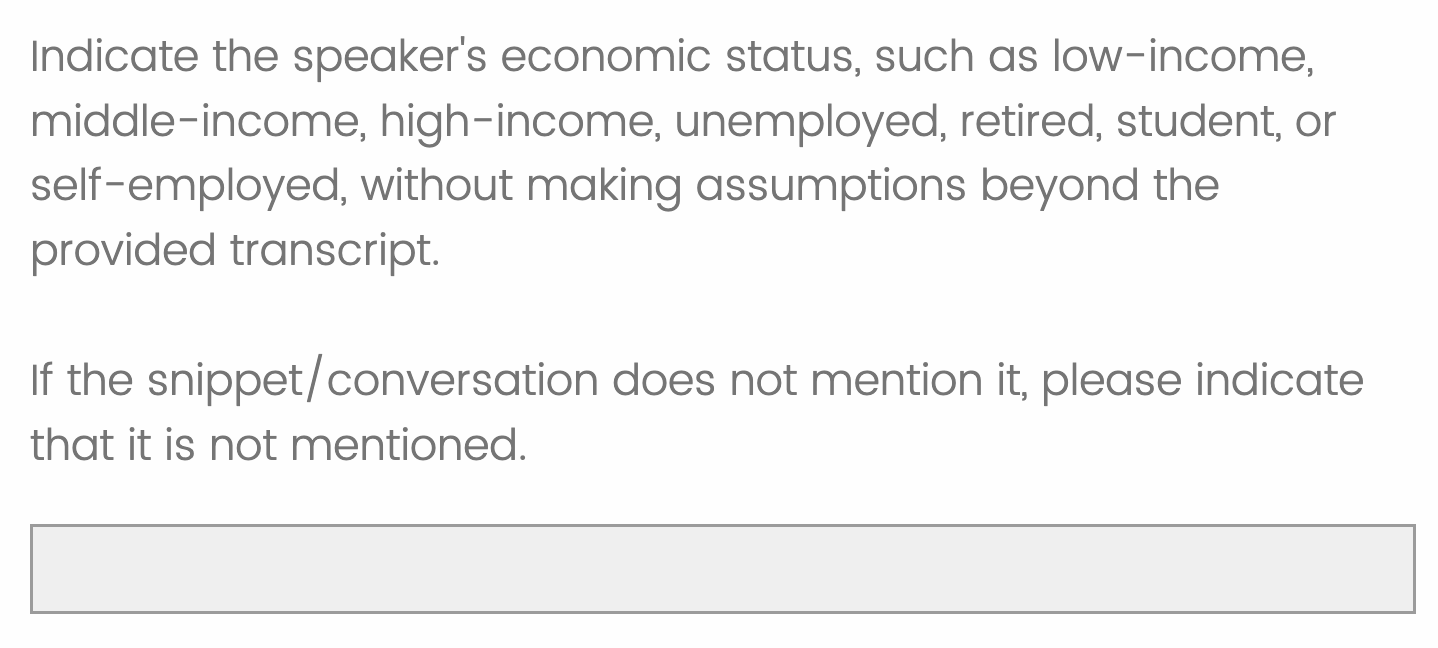}
    \caption{Speaker Economic Status Text Entry Question}
    \label{Figure ##: Speaker Economic Status Text Entry Question}
\end{figure}
\begin{figure}[H]
    \centering
    \includegraphics[width=.8\textwidth]{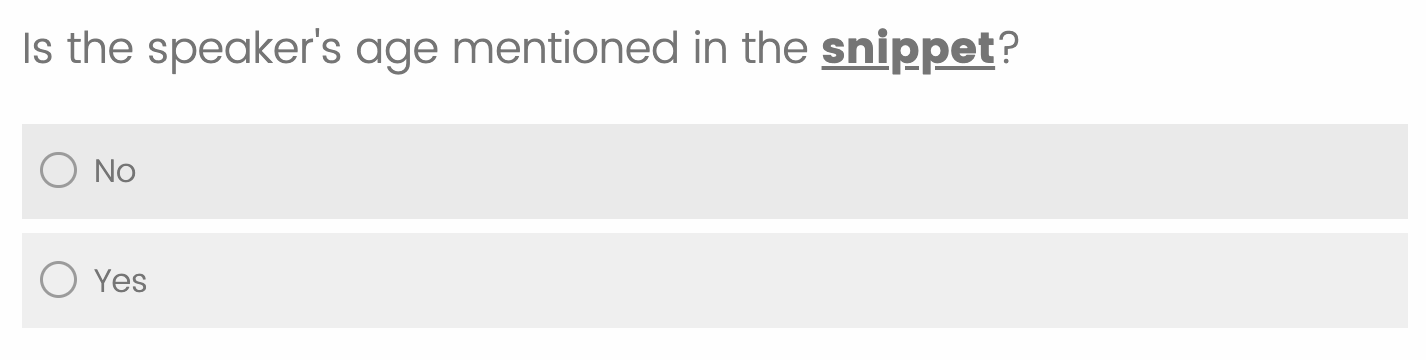}
    \caption{Speaker Age Question}
    \label{Figure ##: Speaker Age YN Question}
\end{figure}
\begin{figure}[H]
    \centering
    \includegraphics[width=.8\textwidth]{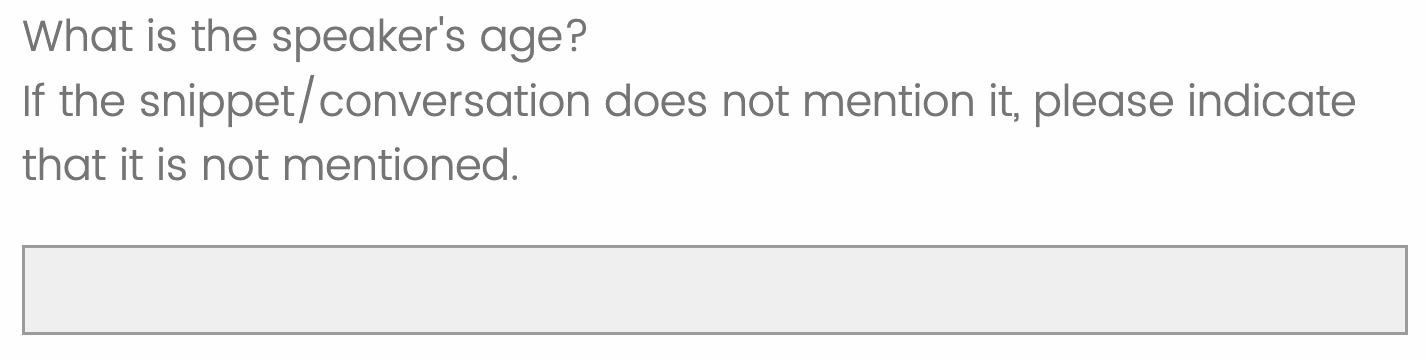}
    \caption{Speaker Age Text Entry Question}
    \label{Figure ##: Speaker Age Text Entry Question}
\end{figure}
\begin{figure}[H]
    \centering
    \includegraphics[width=.8\textwidth]{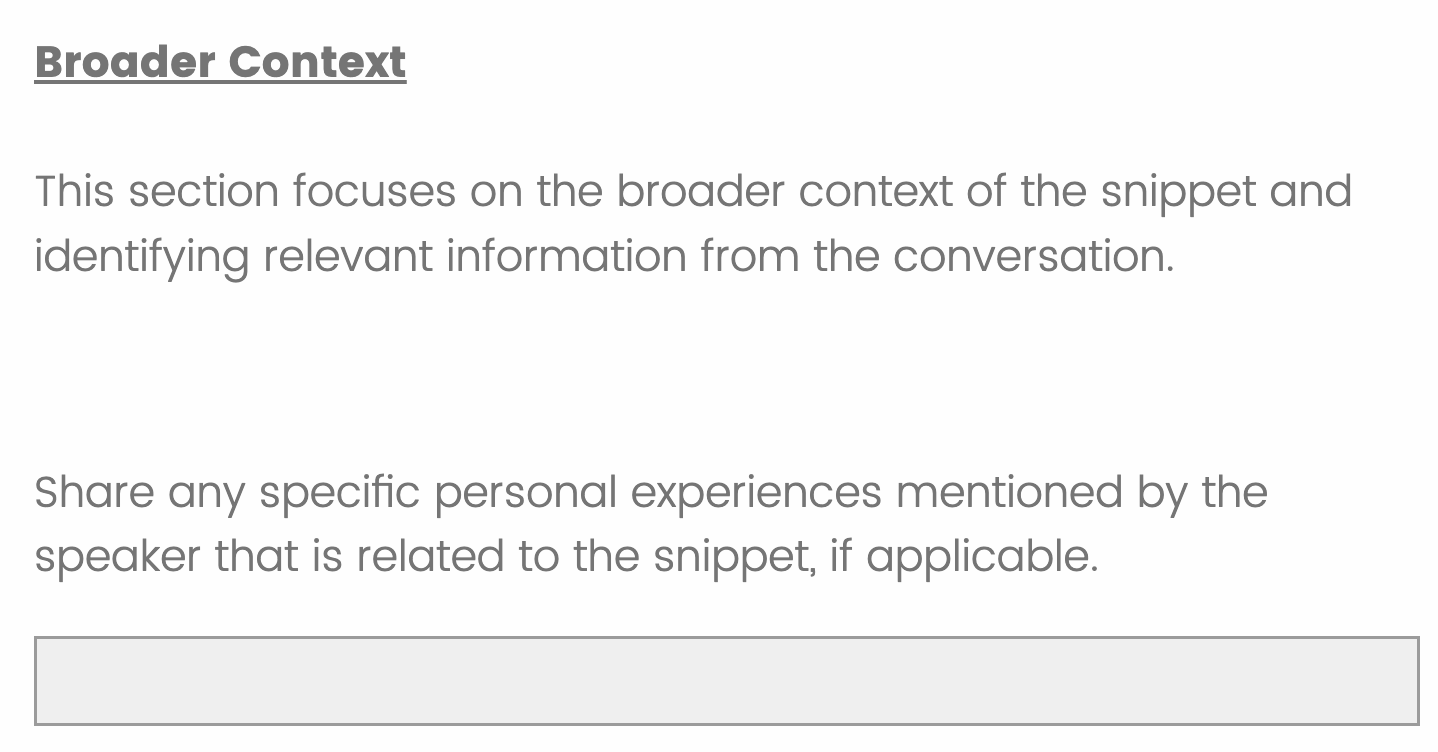}
    \caption{Personal Experience Text Entry Question}
    \label{Figure ##: Personal Experience Text Entry Question}
\end{figure}
\begin{figure}[H]
    \centering
    \includegraphics[width=.8\textwidth]{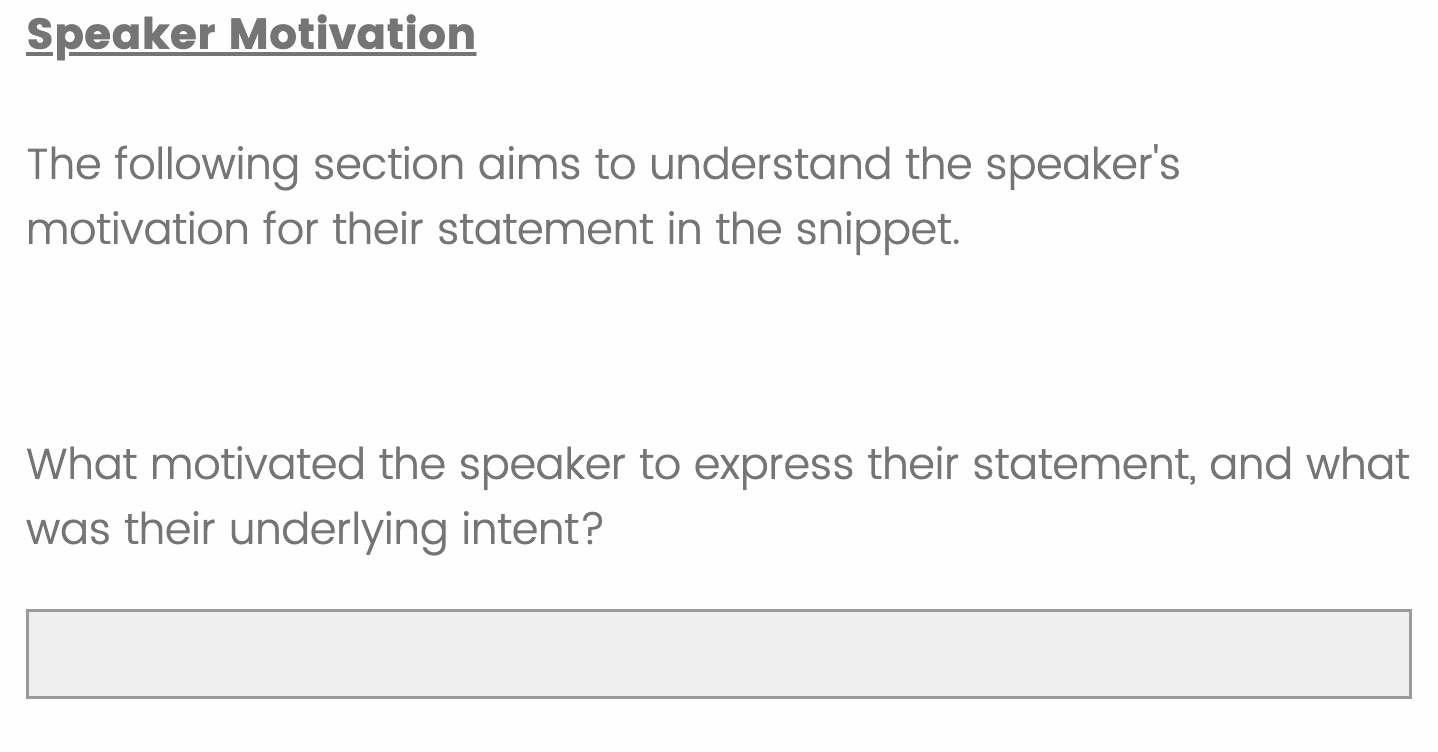}
    \caption{Motivation Text Entry Question}
    \label{Figure ##: Motivation Text Entry Question}
\end{figure}
\begin{figure}[H]
    \centering
    \includegraphics[width=.8\textwidth]{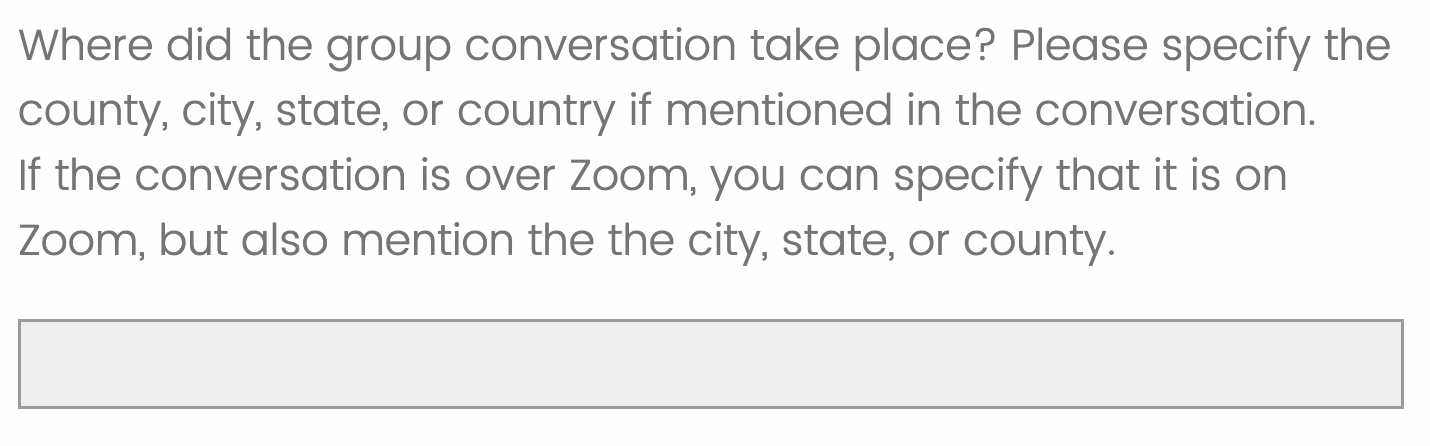}
    \caption{Location Text Entry Question}
    \label{Figure ##: Location Text Entry Question}
\end{figure}
\begin{figure}[H]
    \centering
    \includegraphics[width=.8\textwidth]{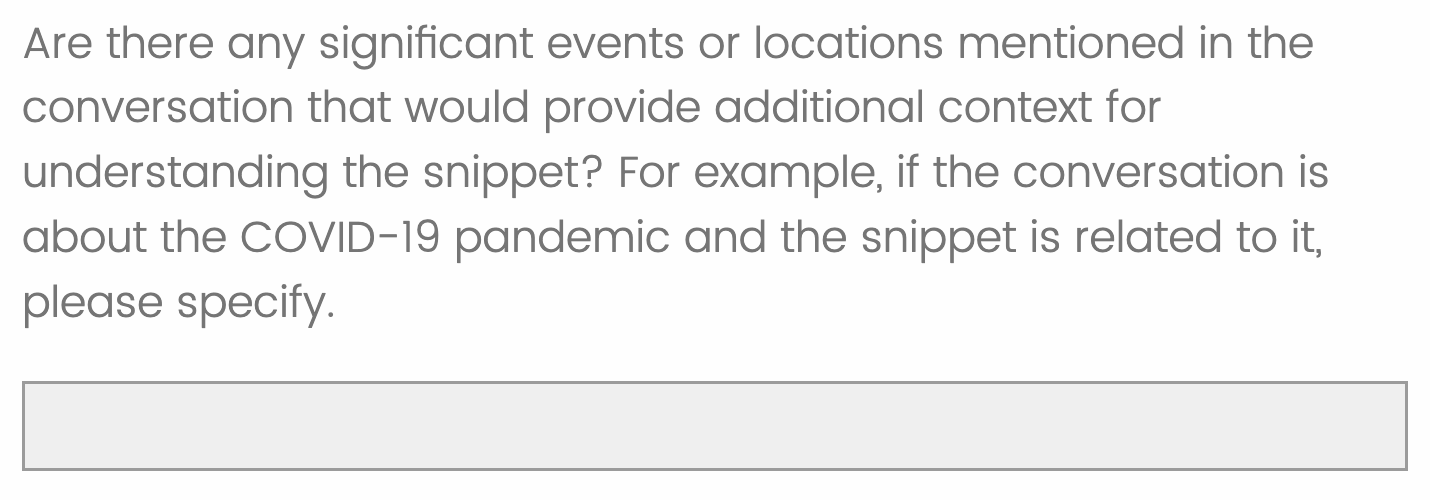}
    \caption{Events Text Entry Question}
    \label{Figure ##: Events Text Entry Question}
\end{figure}
\begin{figure}[H]
    \centering
    \includegraphics[width=.8\textwidth]{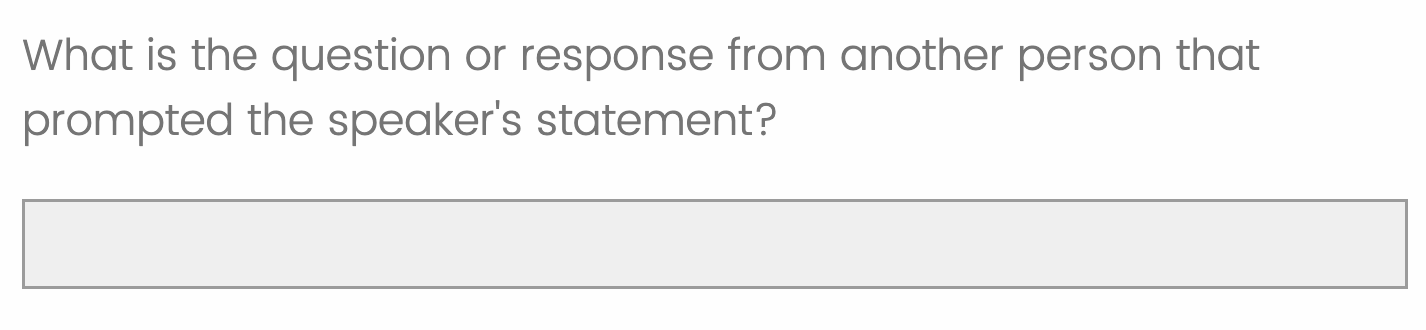}
    \caption{Context Text Entry Question}
    \label{Figure ##: Context Text Entry Question}
\end{figure}
\begin{figure}[H]
    \centering
    \includegraphics[width=.8\textwidth]{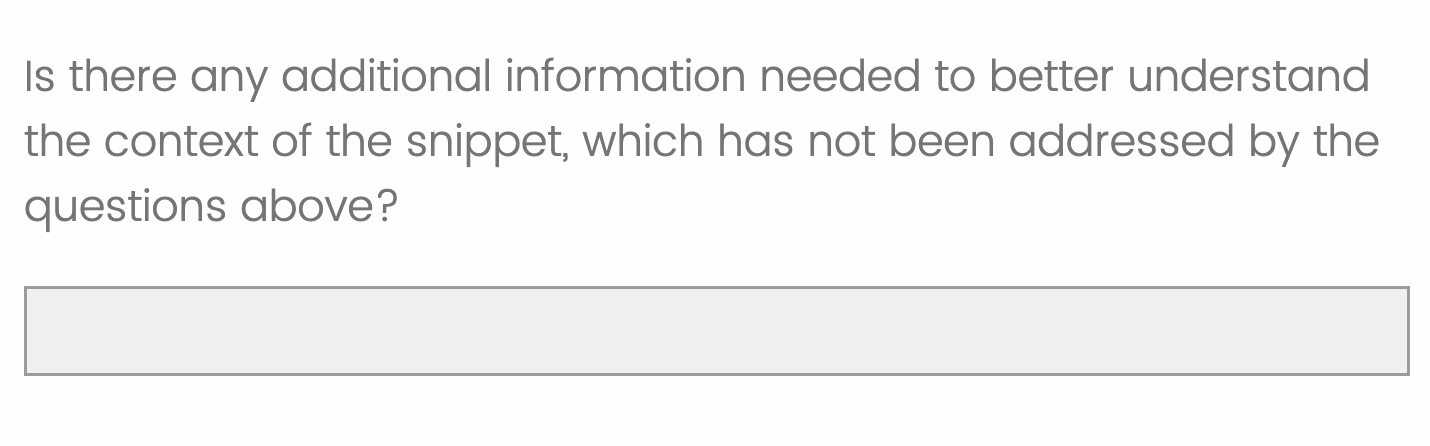}
    \caption{Additional Info Text Entry Question}
    \label{Figure ##: Additional Info Text Entry Question}
\end{figure}

\begin{figure}[H]
    \centering
    \includegraphics[width=.8\textwidth]{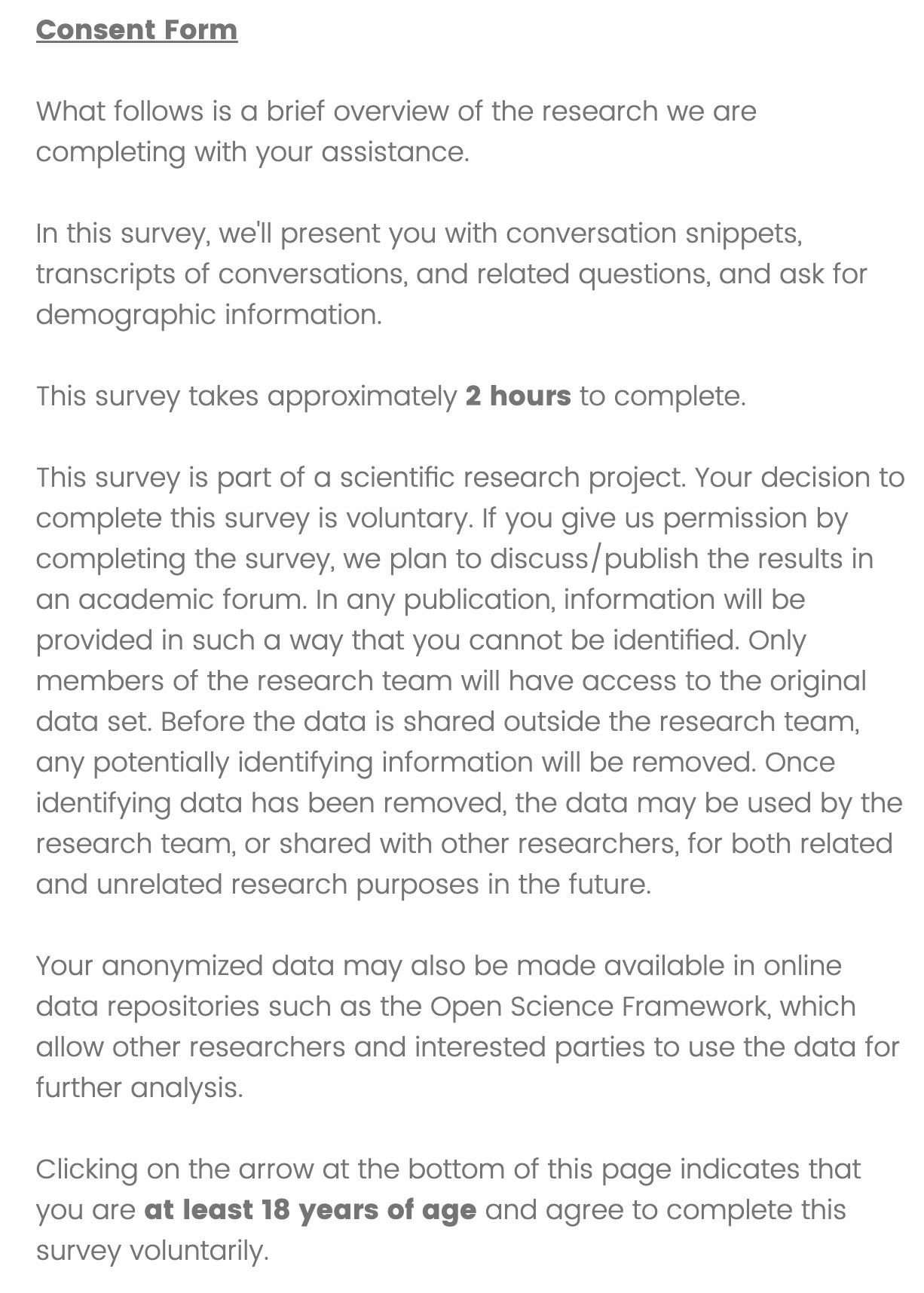}
    \caption{Consent Form}
    \label{Figure ##: Consent Form}
\end{figure}

\end{document}